\documentclass[letterpaper, 10 pt, journal, twoside]{IEEEtran}

\usepackage{booktabs}

\usepackage{graphicx} %
\usepackage{amsmath}
\usepackage{amssymb}
\usepackage{amsfonts}
\usepackage{dsfont}
\usepackage{commath}  %
\usepackage{cuted}  %
\usepackage[hidelinks]{hyperref}
\usepackage[inline]{enumitem}
\usepackage{xcolor}
\usepackage[hang,flushmargin]{footmisc}
\usepackage[export]{adjustbox}  %
\usepackage{tcolorbox} %
\usepackage{pmboxdraw} %
\usepackage{xspace}
\usepackage{flushend}
\usepackage{cite}

\usepackage{subcaption} %
\usepackage[absolute,overlay]{textpos} %

\definecolor{CrimsonRed}{HTML}{DC143C}

\renewcommand{\secref}[1]{Sec.~\ref{#1}}
\renewcommand{\eqref}[1]{Eq.~(\ref{#1})}
\renewcommand{\figref}[1]{Fig.~\ref{#1}}
\newcommand{\tabref}[1]{Tab.~\ref{#1}}

\newcommand\subfiguresubref[1]{Subfig.~(\subref{#1})}
\newcommand\subfiguresubrefbrief[1]{(\subref{#1})}

\newcommand{\etal}{\emph{et al.}}

\newcommand{\mani}{\mathcal{M}}

\newcommand{\ourmethod}{MSG\xspace}
\newcommand{\ourmethodfull}{Multi-Stream Generative Policy\xspace}

\usepackage{siunitx}

\usepackage{threeparttable}
\usepackage{tabularx}
\newcolumntype{Y}{>{\centering\arraybackslash}X}
\newcolumntype{Z}{>{\raggedleft\arraybackslash}X}
\newcolumntype{R}{>{\raggedright\arraybackslash}X}

\newcommand{\flowVariable}{\boldsymbol{z}}

\newcommand{\velocity}{\boldsymbol{v}}

\newcommand{\modelVelocity}{\velocity_\theta}

\newcommand{\pose}{\boldsymbol{\xi}}
\newcommand{\quaternion}{\boldsymbol{q}}
\newcommand{\ee}{\mathrm{ee}}
\newcommand{\sample}{\hat{\pose}}

\newcommand{\rebuttal}[1]{{#1}}

\usepackage{multirow}
\usepackage{makecell}
\usepackage{microtype}

\usepackage{utfsym}
\usepackage{cite}

\usepackage{pifont}%

\usepackage{layouts}  %

\title{
\LARGE \bf
MSG: Multi-Stream Generative Policies for Sample-Efficient\\\vspace{0.5ex}Robotic Manipulation
}

\author{
Jan Ole von Hartz$^{1, *}$, Lukas Schweizer$^{1, *}$, Joschka Boedecker$^{1}$, and Abhinav Valada$^{1}$%
\thanks{This work was supported by Carl Zeiss Foundation with the ReScaLe project, the German Research Foundation (DFG): 417962828, by an academic grant from NVIDIA, and by the BrainLinks-BrainTools center of the University of Freiburg.}
\thanks{$^{1}$Jan Ole von Hartz, Lukas Schweizer, Joschka Boedecker and Abhinav Valada are with the Department of Computer Science, University of Freiburg, Germany. {\tt\footnotesize hartzj@cs.uni-freiburg.de} $^*$Equal contribution.}%
}
\date{March 2026}

\begin{document}

\maketitle

\begin{abstract}
Generative robot policies such as Flow Matching offer flexible, multi-modal policy learning but are sample-inefficient.
Although object-centric policies improve sample efficiency, it does not resolve this limitation. 
In this work, we propose Multi-Stream Generative Policy (MSG), an inference-time composition framework that trains multiple object-centric policies and combines them at inference to improve generalization and sample efficiency. MSG is model-agnostic and inference-only, hence widely applicable to various generative policies and training paradigms. We perform extensive experiments both in simulation and on a real robot, demonstrating that our approach learns high-quality generative policies from as few as five demonstrations, resulting in a 95\% reduction in demonstrations, and improves policy performance by 89 percent compared to single-stream approaches. Furthermore, we present comprehensive ablation studies on various composition strategies and provide practical recommendations for deployment.
Finally, MSG enables zero-shot object instance transfer.
We make our code publicly available at \url{https://msg.cs.uni-freiburg.de}.
\end{abstract}

\section{Introduction}
Generative methods such as Diffusion~\cite{chi2023diffusionpolicy} and Flow Matching~\cite{chisari2024flowmatch} are widely used for robot policy learning as they scale effectively and can represent complex multimodal behaviors.
Their primary limitation is sample inefficiency: they often require on the order of 100 demonstrations to achieve strong performance, whereas probabilistic approaches such as MiDiGaP~\cite{von2025unreasonable} can solve the same tasks with as few as five demonstrations.
A key reason probabilistic policies can generalize from so few examples is that they structure learning into multiple object-centric streams~\cite{calinon2016tutorial}.
In the \texttt{OpenMicrowave} task shown in \figref{fig:multi_frame}, for example, the policy is modeled in the local coordinate frames of the end-effector \emph{and} the microwave.
\rebuttal{Each of these local models constitutes a \emph{stream}: an independently parameterized, object-centric policy component.}
At inference time, these local models are transformed into the world frame using the observed object poses and composed into a single joint model.
This explicit use of geometry enables efficient generalization from very limited data.
In contrast, FiLM-based conditioning commonly used in generative models~\cite{perez2018film} does not encode this structure and struggles to achieve comparable sample efficiency across diverse tasks~\cite{vonhartz2024art, von2025unreasonable}.

Recent object-centric methods have significantly improved generative policy learning's sample efficiency~\cite{rana2024affordancecentricpolicylearningsample}.
However, current approaches still face key limitations. ACPL~\cite{rana2024affordancecentricpolicylearningsample}, for instance, relies on an oriented object frame rather than the full object frame, discarding critical orientation information. Moreover, existing approaches are also limited to a single stream, leaving the benefits of multi-stream learning unexplored.

\begin{figure}
    \centering
    \includegraphics[width=0.9\linewidth]{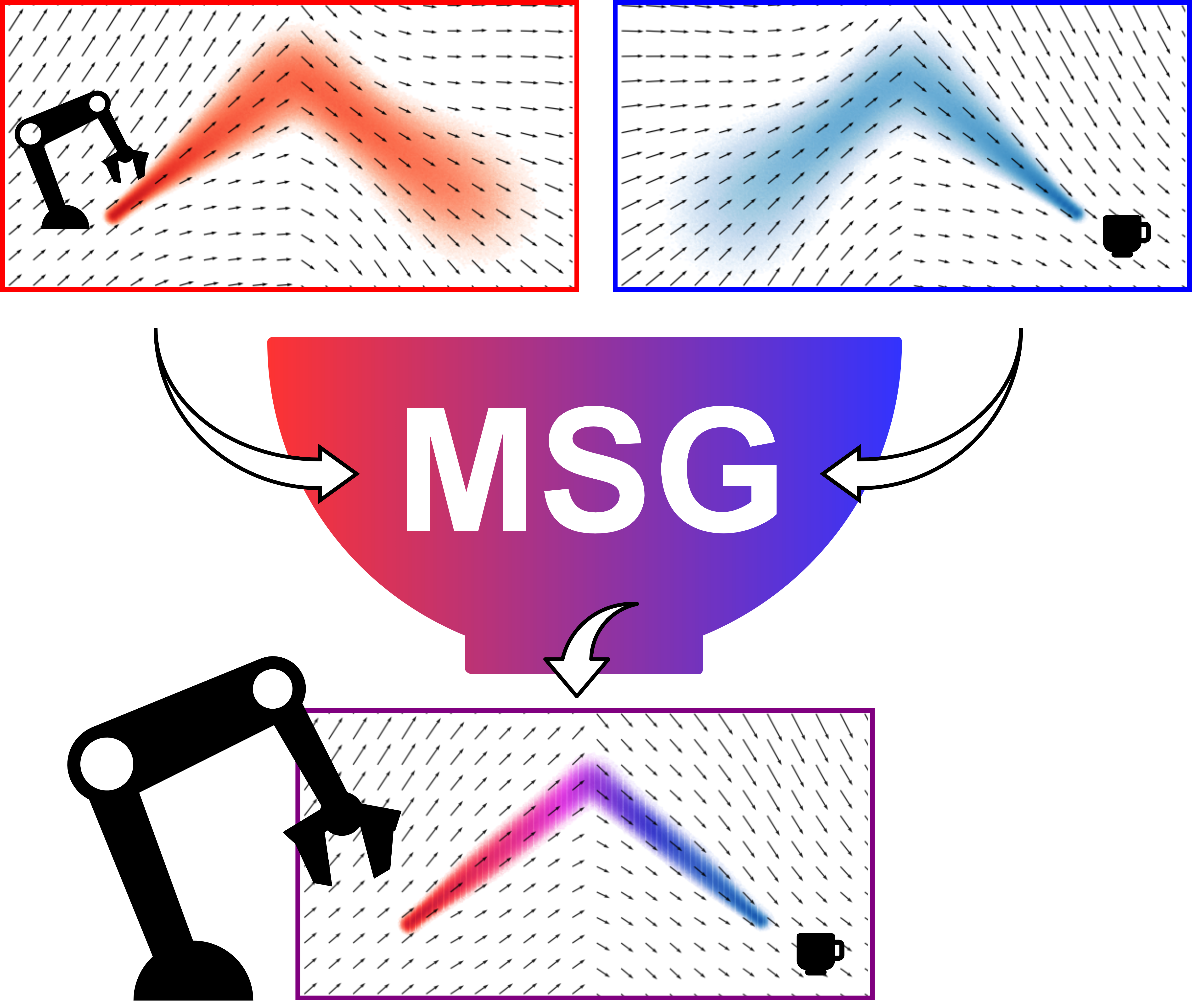}
    \caption{Multi-Stream Generative Policy (MSG) learns high-quality policies from as few as five demonstrations.
    MSG learns multiple object-centric models that are composed at inference time, enabling sample-efficient generalization.}
    \label{fig:eyecatcher}
    \vspace{-6mm}
\end{figure}

We introduce the first multi-stream, object-centric approach for generative policy learning.
Our method, the Multi-Stream Generative Policy (MSG), illustrated in \figref{fig:eyecatcher}, learns multiple object-centric generative policies and composes them at inference. MSG samples from the product of its component models by drawing particles from a shared prior and propagating them through all local flow fields simultaneously.
To integrate the information from these streams, we explore several composition strategies that weight the local models based on their precision, either implicitly or explicitly.

In extensive simulated and real-world experiments, we show that MSG generalizes more effectively from limited data than single-stream approaches. MSG achieves strong performance with as few as five demonstrations, reducing demonstration needs by up to 95\%, and also cuts training time by a similar margin.
Moreover, MSG retains its performance advantage even when scaling to larger datasets.
We evaluate several composition strategies for MSG, each offering a different balance between computational cost, flexibility, and performance.
While our experiments focus on Flow Matching, MSG is model-agnostic and operates only at inference.
It is therefore compatible with other generative policy frameworks and training methods, including Diffusion~\cite{chi2023diffusionpolicy}, Inductive Moment Matching~\cite{zhou2025inductive}, and Implicit Maximum Likelihood Estimation~\cite{rana2025imle}. The inference-only design also enables users to interchange different composition strategies for object-centric policies without retraining.
Building on advances in keypoint estimation~\cite{amir2021deep, vonhartz2024art} and pose estimation~\cite{wen2024foundationpose}, MSG further enables zero-shot generalization across new task environments and object instances.

In summary, our main contributions are the following.
\begin{enumerate}
    \item We propose MSG, a multi-stream object-centric framework for generative policy learning.
    \item We design and evaluate a suite of model composition strategies, offering different trade-offs between inference cost and policy performance.
    \item Through extensive simulated and real-world experiments, we show improved policy generalization, as well as reduced training cost over standard policy learning and existing object-centric approaches.
    \item We release data and source code on our project website.
\end{enumerate}

\section{Related Work}
In this work, we focus on Imitation Learning for robotic manipulation, also known as Learning from Demonstration.
While policies have long been modeled as functions from observation to action~\cite{vonhartz2023treachery, chisari2022correct, florence2019self}, generative approaches have become popular due to their expressivity~\cite{chi2023diffusionpolicy, chisari2024flowmatch}.

{\parskip=2pt
\noindent\textit{Generative Policy Learning}  
estimates the trajectory distribution for solving a given task from a set of demonstrations~\cite{rana2025imle}.
Auto-Encoders~\cite{zhao2023learning} and Energy-Based Models~\cite{florence2022implicit} have been used.
Diffusion~\cite{janner2022planning, chi2023diffusionpolicy, rana2024affordancecentricpolicylearningsample} and Conditional Flow Matching~\cite{chisari2024learningroboticmanipulationpolicies, lipman2022flow, braun2024riemannian} have become popular due to favorable scaling and stable training.
Policy models are typically conditioned on environment observations, such as images~\cite{chi2023diffusionpolicy}, keypoints~\cite{fang2024keypoint, vonhartz2023treachery}, point clouds~\cite{chisari2024learningroboticmanipulationpolicies, Ze2024DP3, 3d_diffuser_actor},  or object poses~\cite{vonhartz2024art, chi2023diffusionpolicy}.
However, such conditioning is less sample-efficient than learning the policy in an object-centric manner~\cite{vonhartz2024art, von2025unreasonable, fang2024keypoint}.}

{\parskip=2pt
\noindent\textit{Object-Centric Policy Learning} models the robot trajectory in a local frame of the object instead of a world frame.
Given an environment observation expressed as an object pose, the model is transformed into the world frame, enabling sample-efficient generalization across the task space.
This approach is favored for probabilistic policies such as Gaussian Mixture Models (GMM)~\cite{calinon2016tutorial, zeestraten2018programming, vonhartz2024art} and Discrete-Time Gaussian Process Mixtures (MiDiGaP)~\cite{von2025unreasonable}, but has recently been applied to generative policies as well~\cite{rana2024affordancecentricpolicylearningsample, fang2024keypoint}.
Notably, Affordance-Centric Policy Learning (ACPL)~\cite{rana2024affordancecentricpolicylearningsample} leverages an oriented object frame that aligns one axis of the frame with the end-effector.
While improving the support of the trajectory distribution, this orientated frame forfeits the object rotation.
In contrast, we argue that a better way to improve policy generalization for generative models is multi-stream learning.}

{\parskip=2pt
\noindent\textit{Multi-Stream Object-Centric Learning} models the trajectory distribution in the local coordinate frames of multiple objects~\cite{calinon2016tutorial}.
This approach is effective for probabilistic methods that can easily compute the product of multiple trajectory distributions, such as Task-Parameterized GMMs~\cite{calinon2016tutorial, zeestraten2018programming, vonhartz2024art} and MiDiGaP~\cite{von2025unreasonable}.
In contrast, the composition of generative policy models is less straightforward.
Patil \etal{} argue that a weighted linear combination of the score functions, predicted by two Diffusion models, is proportional to the gradient of the corresponding weighted mixture of energy functions~\cite{patil2024composing}.
PoCo takes the same approach to the composition of Diffusion Policies\cite{wang2024poco}.
However, Du \etal{} show that summing up the score predictions of two Diffusion models is \emph{not} equivalent to sampling from the product distribution and propose an annealed MCMC schema instead~\cite{du2023reduce}.
Wang \etal{} applied this principle to steering Diffusion policy with a point objective function~\cite{wang2024inference}.
Such compositional approaches have not yet been adapted to Flow Matching policies.}

To the best of our knowledge, we are the first to employ a multi-stream object-centric approach to generative policies.
As we show in extensive experiments, multi-stream learning consistently improves policy success both over naive policy learning and over single-stream approaches, such as ACPL.
Using comprehensive ablation studies, we investigate policy composition strategies for Conditional Flow Matching.

\section{Background}\label{sec:back}
\subsection{Generative Policy Learning}
Given a dataset \(\mathcal{D}=\{\left(\boldsymbol o_s^n, \boldsymbol a_s^n\right)_{s=1}^{T_n}\}_{n=1}^N\) of observations \(\boldsymbol o\) and actions \(\boldsymbol a\), imitation learning aims to maximize the likelihood of observed actions given observations, i.e.,\
\begin{equation}
\theta^* = \arg\max_\theta \; \mathbb{E}_{(\boldsymbol o, \boldsymbol a) \sim D} \left[ \log \pi_\theta(\boldsymbol a \mid\boldsymbol  o) \right].
\end{equation}
Instead of learning the mapping from observations to actions directly, Conditional Flow Matching poses policy learning as an ordinary differential equation (ODE) of the form 
\begin{equation}\label{eq:ode}
    \frac{\mathrm{d}}{\mathrm{d}t} \flowVariable_t = \velocity_{\boldsymbol\theta}(\boldsymbol\flowVariable_t, t\mid \boldsymbol o), \qquad \flowVariable_0 \sim p_0,
\end{equation}
with an arbitrary prior distribution $p_0$ - usually a standard normal Gaussian.
The objective is to learn the vector field \( \velocity_{\boldsymbol\theta}: \mathbb{R}^d \times[0,1] \rightarrow \mathbb{R}^d \) that transports the prior $p_0$ to the target distribution $p_1$ observed under the dataset \(\mathcal{D}\), in unit time.
The neural network used to model \( \velocity_{\boldsymbol\theta}\) is learned by optimizing

{\small
\begin{equation} \label{eq:loss}
\min_{\boldsymbol\theta} \mathbb{E}_{ \flowVariable_0 \sim p_0, (\boldsymbol o, \boldsymbol a) \sim p_1 }
    \left[
        \int_0^1 \left\| 
            \boldsymbol a - \flowVariable_0
            -
            \modelVelocity\left(\flowVariable_t, t\mid\boldsymbol o\right)
            \right \|_2^2 \mathrm{~d} t
    \right],
\end{equation}
}
where \(\boldsymbol z_t=t \boldsymbol z_0 + (1-t) \boldsymbol z_1\) linearly interpolates \(\boldsymbol z_0\) and \(\boldsymbol a=\boldsymbol z_1\)~\cite{liu2023flow,lipman2023flow}.
For inference, we sample \(\flowVariable_0\sim p_0\) and numerically integrate the ODE in \eqref{eq:ode} up to time \(t=1\).
Note that we model the pose \(\pose=\begin{bmatrix}\boldsymbol x & \quaternion\end{bmatrix}^T\) of the robot's end-effector on the manifold \(\mani_{\text{pose}}=\mathbb{R}^3\times\mathcal{S}^3\).
To this end, we extend \eqref{eq:loss} to match geodesic flows on \(\mani_{\text{pose}}\)~\cite{lipman2024flow}, ensuring that intermediate samples  \(\boldsymbol z_t\) are valid poses.

\subsection{Object-Centric Policy Learning}\label{sec:back_obj_centric}
Generative policies usually model the \emph{absolute} pose \(\pose_\ee\) of the robot's end-effector.
For instance, Diffusion Policy predicts the next \(T_p\) end-effector poses given the last \(T_o\) observations~\cite{chi2023diffusionpolicy}.
Such naive conditioning leads to sample-inefficient learning, for it requires the model to learn the equivariance relationship between object poses and end-effector pose~\cite{vonhartz2024art}. 
If object poses are available or can be estimated, we can instead model the end-effector pose in the local frame of a task-relevant object, thus simplifying the learning problem.

Given a local coordinate frame \(f\) with pose \(\pose_f=[\boldsymbol x_f\ \ \quaternion_f]^T\), the end-effector pose \(\pose_\ee^{(f)}\) in the local frame \(f\) is given by
\begin{equation}\label{eq:frametrans}
    \pose_\ee^{(f)} = \begin{bmatrix}
    \quaternion_f^{-1} (\boldsymbol x_\ee  - \boldsymbol x _f) \quaternion_f & 
    \quaternion_f^{-1} \quaternion_\ee
    \end{bmatrix}^T.
\end{equation}
The policy then models the density \(p\left(\pose_\ee^{(f)}\right)\).
During inference, the global pose \(\pose_\ee\) is reconstructed by inverting \eqref{eq:frametrans}.

{\parskip=2pt
\noindent\textit{Skill Chaining:}
To solve long-horizon tasks involving multiple objects,  multiple local skill models are learned and sequenced~\cite{vonhartz2024art}.
For generative policy models, one annotates a demonstration trajectory \((\pose_s)_{s=1}^S\) with a \emph{progress} value \(p_t=\frac{s}{S}\) and trains the policy to jointly predict both action and progress value~\cite{rana2024affordancecentricpolicylearningsample}.\footnote{Note that \(s\) refers to the time step of the pose in the \emph{robot trajectory}, while \(t\) is reserved for the flow step. We drop other subscripts here for lucidity.}
During inference, a skill transition is performed when the predicted progress value exceeds a given threshold.}

{\parskip=2pt
\noindent\textit{Oriented Object Frames:}
While object-centric learning effectively generalizes across the task space, it does not take the robot embodiment into account~\cite{von2025unreasonable}.
Consequently, Rana~\etal{} report frequent joint limit violations when applying object-centric learning to Diffusion Policy~\cite{rana2024affordancecentricpolicylearningsample}.
As a remedy, they propose to orient the x-axis of the object frame towards the end-effector, thus stabilizing the end-effector orientation~\cite{rana2024affordancecentricpolicylearningsample}.
While reducing joint limit violations, orienting the object frame in this way effectively discards orientation information captured by the local model, such as the approach angle needed to grasp an object.
As we show in \secref{sec:rlbench} using standardized benchmarks, orienting the object frame usually \emph{affects} policy performance.
Instead, we propose to leverage multi-stream learning to stabilize the prediction.}

\begin{figure}
    \centering
    \includegraphics[width=0.66\linewidth]{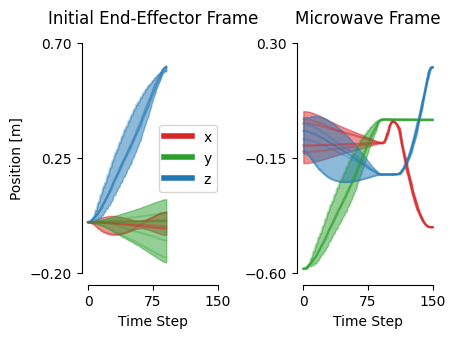}%
    \includegraphics[width=0.33\linewidth]{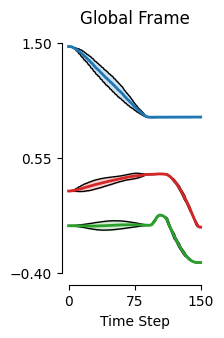}
    \caption{Multi-stream model for \texttt{OpenMicrowave}, plotting the three position dimensions over time.
    The local model in the initial end-effector frame has low variance at the beginning of the trajectory and becomes less informative over time.
    In contrast, the microwave's local model gets more informative as the end-effector approaches the microwave.
    Once the end-effector grasps the handle, the end-effector frame is dropped entirely as it no longer adds information.
    The combined model has high precision throughout the trajectory.
    }
    \label{fig:multi_frame}
    \vspace{-3mm}
\end{figure}

\subsection{Multi-Stream Learning}\label{sec:multi_stream}
Given a set of \(F\) local coordinate frames, multi-stream methods estimate  \(F\) object-centric models \(\{p(\pose_\ee^{(f)})\}_{f=1}^F\)~\cite{calinon2016tutorial}.
During inference, these local models are transformed into the world frame, yielding \(\{p(\pose_\ee\mid f)\}_{f=1}^F\).
Assuming conditional independence of \(\{f\}_{f=1}^F\), they are integrated into a joint model via\looseness=-1
\begin{equation}\label{eq:comb}
   p\left(\pose_\ee\mid \{f\}_{f=1}^F\right) \propto \prod_{f=1}^F p(\pose_\ee\mid f).
\end{equation}
For Gaussian policies, this product is implemented via the product of Gaussians.
In \secref{sec:approach}, we discuss the application to generative policies.
\figref{fig:multi_frame} illustrates a Gaussian multi-stream policy for \texttt{OpenMicrowave}~\cite{von2025unreasonable}.
While the local model in the initial end-effector frame has low variance in the beginning of the trajectory, the microwave's local model is more informative later in the trajectory.
The combined model in the world frame thus has low variance across the whole trajectory.
Crucially, without access to ground-truth object poses, the task-relevant frames can be estimated and automatically selected from image observations using DINO keypoints and stochastic considerations~\cite{vonhartz2024art}.
By leveraging DINO keypoints~\cite{amir2021deep}, these methods further generalize across task environments and object instances~\cite{vonhartz2024art, von2025unreasonable}.

\section{Technical Approach}\label{sec:approach}
Our goal is to learn generative policies from few samples.
For probabilistic policies, multi-stream object-centric learning has proven effective.
These methods train multiple independent models, or streams, which are combined only at inference using a product-of-experts formulation.
This separation of training and composition is especially appealing in generative policy learning, a rapidly evolving field where new training techniques continue to emerge~\cite{zhou2025inductive, rana2025imle}.
Recent work also highlights the constitutionality of generative policies~\cite{du2023reduce}.
Building on these ideas, we propose a method for applying multi-stream learning to generative policies.

\begin{figure*}[ht]
    \centering
    \captionsetup[subfigure]{justification=centering}
    \begin{subfigure}[t]{0.5\textwidth}
        \centering
        \includegraphics[height=3.6cm, trim={0cm 0cm 12.8333cm 0cm}, clip]{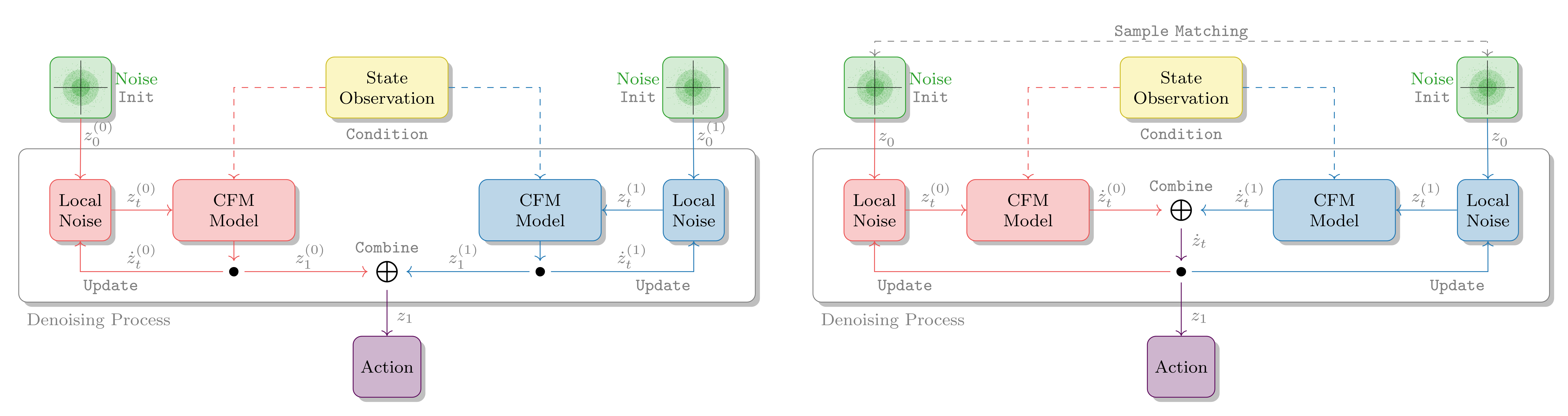}
        \caption{The \emph{ensemble} fully integrates the individual streams before composing them.}\label{fig:comb_ens}
    \end{subfigure}%
    \begin{subfigure}[t]{0.5\textwidth}
        \centering
        \includegraphics[height=3.6cm, trim={12.8333cm 0cm 0cm 0cm}, clip]{figures/combined_inference_flat_v2_aligned.png}
        \caption{The \emph{flow composition} jointly integrates the flow fields, combining them at each step.}\label{fig:comb_icss}
    \end{subfigure}
    \caption{We compare \subfiguresubrefbrief{fig:comb_ens} composing the fully integrated predictions of the local generative models with \subfiguresubrefbrief{fig:comb_icss} iteratively composing their predicted vector fields.}
    \vspace{-3mm}
\end{figure*}

\subsection{Learning Generative Multi-Stream Policies}\label{sec:ms_flow_learn}
As discussed in \secref{sec:back_obj_centric} and \secref{sec:multi_stream}, multi-stream learning models global end-effector trajectories in a set of local coordinate frames.
Given a dataset \(\mathcal{D}\) of global end-effector poses  \(\pose_\ee\) and frame poses \(\{\pose_f\}_{f=1}^F\), we transform the end-effector poses into each local coordinate frame \(f\), yielding datasets \(\{\mathcal{D}^{(f)}\}_{f=1}^F\) of local poses \(\pose_\ee^{(f)}\).
On each dataset \(\mathcal{D}^{(f)}\), we train a generative policy \(\velocity_f\)  modeling the local density \(p(\pose_\ee^{(f)})\).
I.e.\ the individual streams are trained on ODE dynamics that are equivalent up to frame transformation of the reference paths.
This alignment is critical to ensure proper composition of the models.
In our experiments, we train a Riemannian Flow Matching policy~\cite{lipman2023flow}. Nevertheless, our method is applicable to other generative approaches, such as Diffusion~\cite{chi2023diffusionpolicy}, Inductive Moment Matching~\cite{zhou2025inductive}, and Implicit Maximum Likelihood Estimation~\cite{rana2025imle}.
For the most part, we follow the standard procedure for training Flow Matching described in \secref{sec:back}.
However, as we explain in \secref{sec:model_comp}, we construct a custom prior \(p_0\) in some cases.

\subsection{Inference-Time Model Composition}\label{sec:model_comp}
Having learned multiple, conditionally independent object-centric models \(\{\velocity_f\}_{f=1}^F\), we face two main challenges during inference.
First, we must transform the local models \(\{\velocity_f\}_{f=1}^F\), which define local densities \(\{p(\pose_\ee^{(f)})\}_{f=1}^F\), into the common world frame to obtain the global densities \(\{p(\pose_\ee\mid f)\}_{f=1}^F\).
Second, we need to estimate their product distribution \( p(\pose_\ee\mid \{f\}_{f=1}^F)\), as defined in \eqref{eq:comb}.
For generative policies like Flow Matching, both steps are intractable for these models represent trajectory distributions implicitly, not in closed form.
However, while we cannot directly transform the implicit distribution \(p(\pose_\ee^{(f)})\), we can easily transform samples drawn from it.
Since our goal is to \emph{sample} from the composite distribution, a sample-based approximation of both operations becomes a practical and effective solution.
We explore two composition strategies: an ensemble-based approach and one based on flow composition.

{\parskip=2pt
\noindent\textit{Ensemble-Based Composition:} 
A simple way to approximate sampling from the joint distribution is through a single-particle, ensemble-based method, as illustrated in \figref{fig:comb_ens}.
To this end, we draw a single sample \(\flowVariable_0 \sim p_0\) and integrate it independently through each of the learned flows \(\{\velocity_f\}_{f=1}^F\).
This yields local-frame samples}
\begin{equation}
    \left\{\sample_\ee^{(f)}\sim p\left(\pose_\ee^{(f)}\right)\right\}_{f=1}^F.
\end{equation}
We transform these samples into the world frame, resulting in
\begin{equation}
    \{\sample_{\ee, f}\sim p(\pose_\ee\mid f)\}_{f=1}^F
\end{equation}
and perform a weighted combination of the samples as
\begin{equation}\label{eq:ens}
    \sample_\ee\approx\sum_{f=1}^F \rebuttal{\boldsymbol w_f \odot \sample_{\ee, f}}.
\end{equation}
Written in Euclidean form for clarity, interpolation is performed geodesically on the pose manifold.
\rebuttal{\(\odot\) denotes elementwise multiplication.
The weights \(\boldsymbol w_f\) may be scalar or vector-valued (with broadcasting as needed); see \secref{sec:weight} for details.}

\begin{figure}[t]
    \centering
    \includegraphics[width=\linewidth]{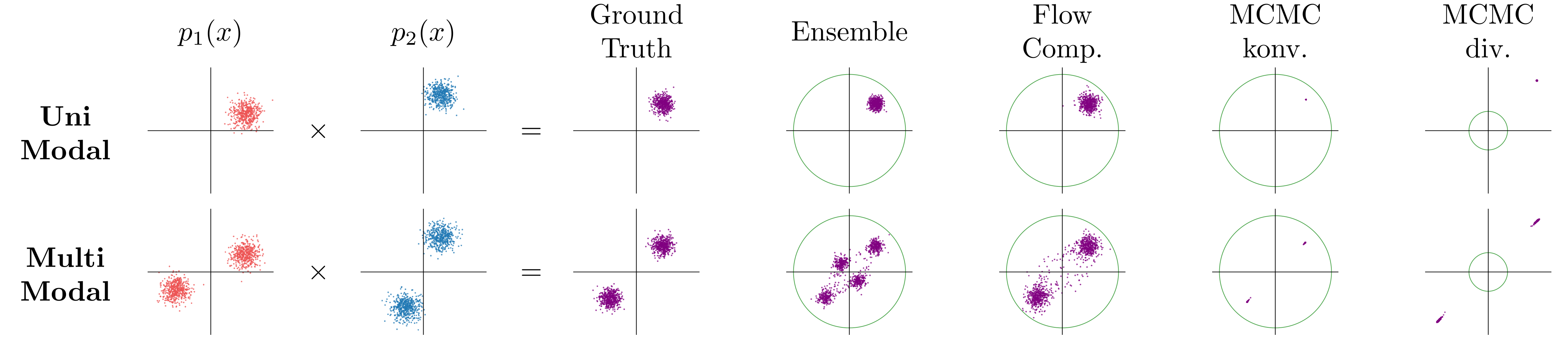}
    \caption{Stream composition strategies.
    The green circles indicate the one-sigma interval of the zero-mean Gaussian prior.
    For unimodal targets, the ensemble approximates the product distribution well.
    For multimodal targets, it fails because it lacks a mechanism to guide both streams to the same mode.
    Flow Composition encourages convergence to a common mode, reinforced by MCMC.
    MCMC can underestimate the target variance or overshoot if the prior is unlikely under the target; a suitable prior ensures converging flows.
    }
    \label{fig:toy_comp}
    \vspace{-5mm}
\end{figure}

For general distributions, \eqref{eq:ens} does not yield a valid approximation of a sample drawn from the product of distributions.
The approximation is only meaningful if both distributions are close to being Gaussian.
However, recent work has shown that unimodal, object-centric trajectory distributions can be successfully approximated by Gaussian policies~\cite{vonhartz2024art, von2025unreasonable}.
Furthermore, while we focus on unimodal distributions in this work,  \emph{mixtures} of Gaussian policies can represent multimodal distributions~\cite{von2025unreasonable}.
In the multimodal case, the key challenge for ensemble-based composition is ensuring that all samples come from the same mode.
As illustrated in \figref{fig:toy_comp}, this alignment is more reliably achieved by composing the streams' flows directly, rather than combining their final predictions.

{\parskip=2pt
\noindent\textit{Flow Composition:} 
For diffusion models, the product-of-experts composition in \eqref{eq:comb} has been approximated by summing the scores predicted by the individual models~\cite{liu2022compositional, wang2024poco}.
However, Du \etal{} show that this approach does not truly sample from the joint distribution~\cite{du2023reduce}. Instead, they propose using annealed MCMC, performing multiple reverse process steps at each noise level.
We adapt this procedure to flow matching, as illustrated in \figref{fig:comb_icss}.
Unlike diffusion models, flow matching does not predict a score function but rather a time-dependent vector field~\cite{lipman2022flow}.
Strictly speaking, this kind of composition is only valid if the vector field is gradient-like.
Nevertheless, our experiments show that composing flows in this way performs well in practice, with only minor caveats.}\footnote{
As \figref{fig:toy_comp} shows, non-gradient-like vector fields cause two types of MCMC approximation errors.
First, underestimation of the target distribution's variance.
In policy learning, we care primarily about sample quality, not diversity, so this is a minor issue.
Second, if the target lies far from the prior, the straight flows can overshoot it.
Diffusion avoids this by approximating the score function and following curved denoising paths, ensuring particle convergence.
We mitigate overshooting by ensuring sufficient likelihood of the prior under the target distribution.
}

We again draw a single sample \(\flowVariable_0 \sim p_0\) in the global frame, and transform it into the local frame for each model, yielding \(\{\flowVariable_0^{(f)}\}_{f=1}^F\).
During each flow time step \(t\) and for each local model \(\velocity_f\), we infer \(\dot{\flowVariable}_{t}^{(f)} = \velocity_f(\flowVariable_t^{(f)})\).
These local vectors are transformed to world frame, yielding \(\{\dot{\flowVariable}_{t,f}\}_{f=1}^F\), and combined as
\begin{equation}\label{eq:flow_comp}
    \dot{\flowVariable}_{t}\approx\sum_{f=1}^F \rebuttal{\boldsymbol w_{t, f}\odot \dot{\flowVariable}_{t,f}}.
\end{equation}
The joint flow \(\dot{\flowVariable}_{t}\) is then used to update \({\flowVariable}_{t}\) as usual.
Again, we discuss the employed weighting strategies in \secref{sec:weight}.

\begin{figure*}[tb]
    \centering
    \includegraphics[width=0.071\textwidth,valign=t]{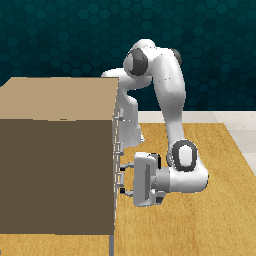}\hfil
    \includegraphics[width=0.071\textwidth,valign=t]{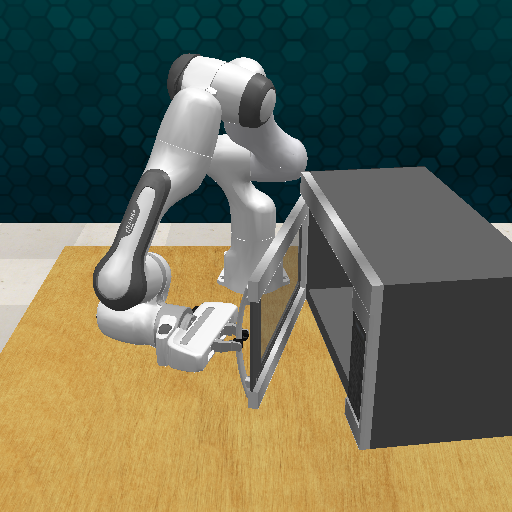}\hfil
    \includegraphics[width=0.071\textwidth,valign=t]{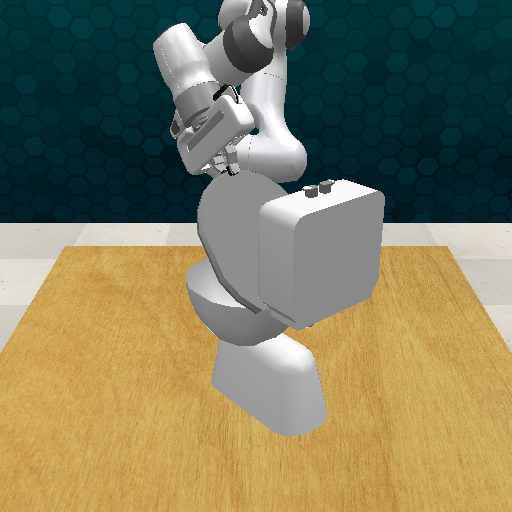}\hfil
    \includegraphics[width=0.071\textwidth,valign=t]{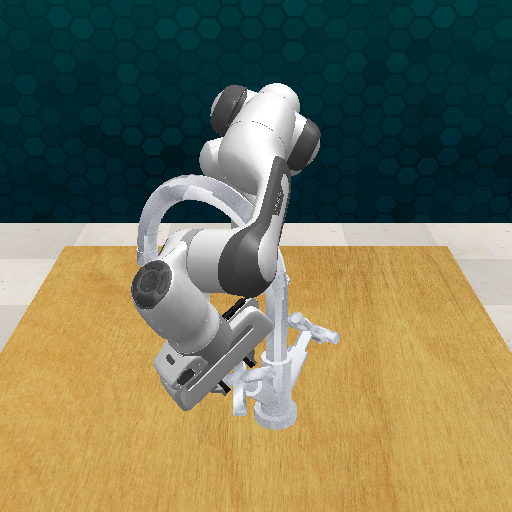}\hfil
    \includegraphics[width=0.071\textwidth,valign=t]{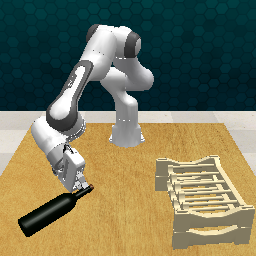}\hfil
    \includegraphics[width=0.071\textwidth,valign=t]{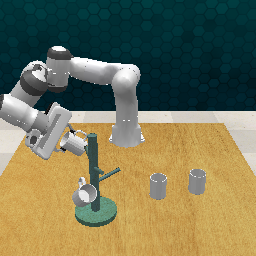}\hfil
    \includegraphics[width=0.071\textwidth,valign=t]{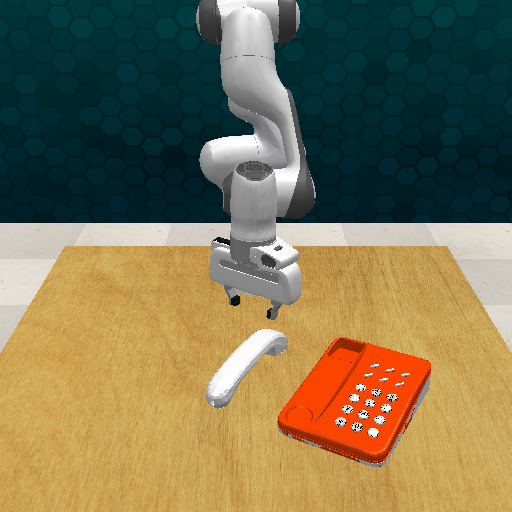}\hfil
    \includegraphics[width=0.071\textwidth,valign=t]{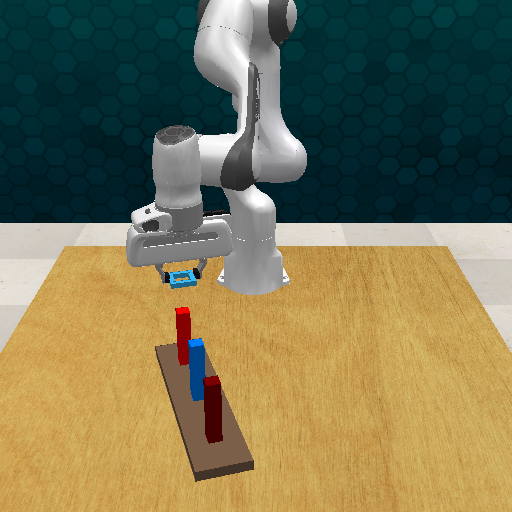}\hfil
    \includegraphics[width=0.071\textwidth, valign=t]{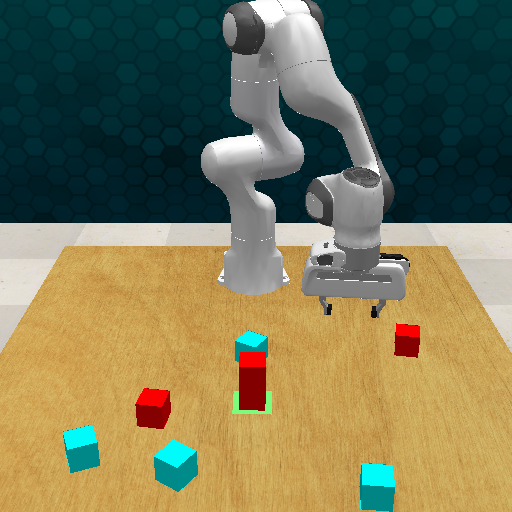}\hfil
    \includegraphics[width=0.071\textwidth,valign=t]{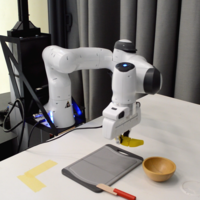}\hfil
    \includegraphics[width=0.071\textwidth,valign=t]{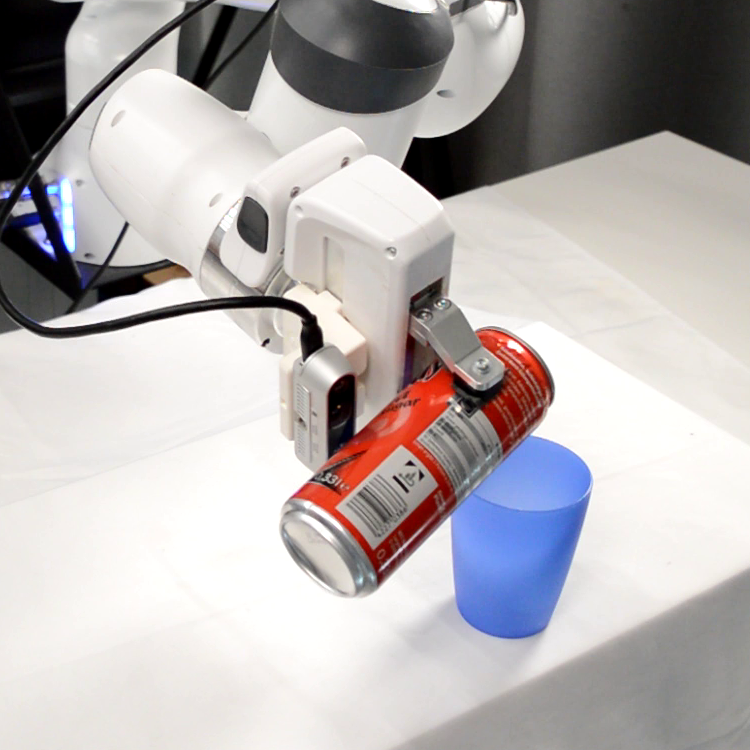}\hfil
    \includegraphics[width=0.071\textwidth,valign=t]{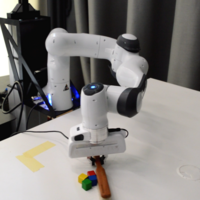}\hfil
    \includegraphics[width=0.071\textwidth,valign=t]{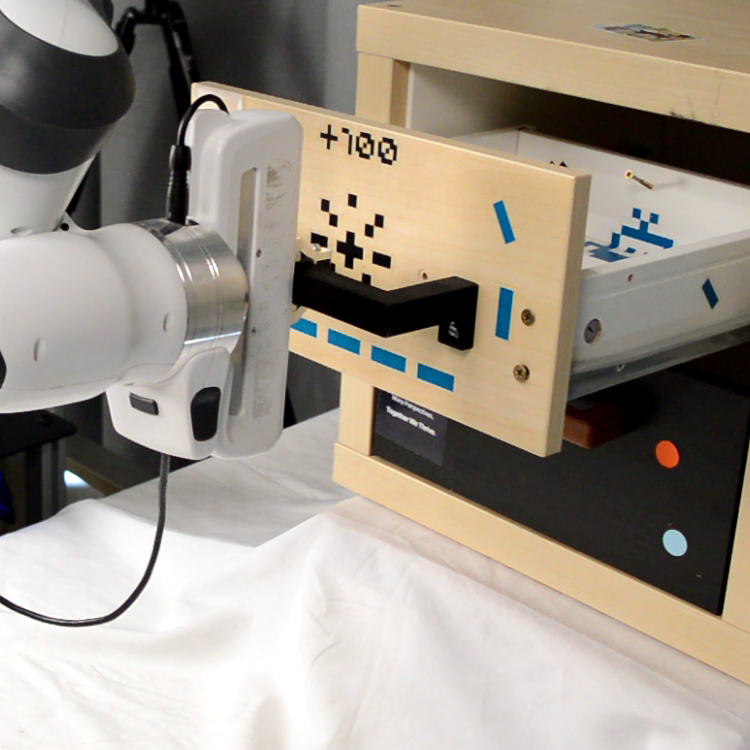}\hfil
    \includegraphics[width=0.071\textwidth,valign=t]{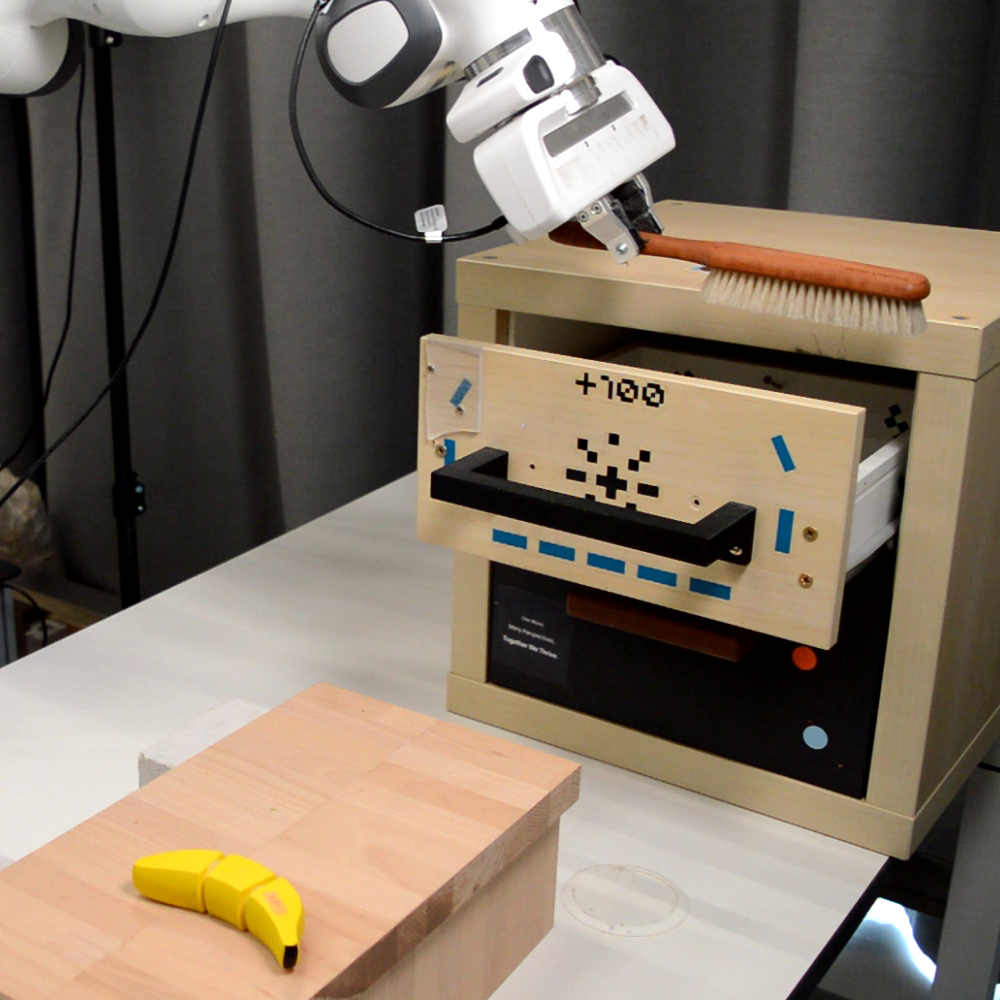}
    \caption{The RLBench tasks \texttt{OpenDrawer}, \texttt{OpenMicrowave}, \texttt{ToiletSeatUp},  \texttt{TurnTap}, \texttt{StackWine},  \texttt{PlaceCups}, \texttt{PhoneOnBase}, \texttt{InsertOntoSquarePeg}, \rebuttal{\texttt{StackBlocks}}. The real-world tasks \texttt{PickAndPlace}, \texttt{PourDrink}, \texttt{SweepBlocks}, \texttt{OpenDrawer}, \rebuttal{and \texttt{StoreInDrawer}}.}\label{fig:tasks}
   \vspace{-0.5cm}
\end{figure*}

Note that we match the initial samples passed through each model by sampling the initial noise in the global frame and transforming it into the local frames.
This accounts for the inverse frame transformation from \(\dot{\flowVariable}_{t}^{(f)}\) to \(\dot{\flowVariable}_{t,f}\) applied before flow composition in \eqref{eq:flow_comp}.
As we show empirically in \secref{sec:exp}, this sample alignment is crucial for effective composition.
However, this sampling procedure means that the initial sample \(\boldsymbol\flowVariable_0^{(f)}\) in each model is no longer drawn from a standard Gaussian.
To prevent distribution shift, we do not train the local models in \secref{sec:ms_flow_learn} with a standard Gaussian prior.
Instead, we construct a custom Gaussian prior centered around the current local end-effector pose \(\pose_\ee^{(f)}\).
Alternatively, we also construct a Gaussian mixture prior by transforming the standard Gaussian into each local frame according to the object pose distribution observed under the demonstration data.
Both priors alleviate distribution shift.
But in contrast to the mixture prior, the pose-centric prior further encourages convergence of the learned vector field, and thus of MCMC.
We verify these hypotheses in \secref{sec:ablation}.

\subsection{Weighting Strategies}\label{sec:weight}
Both composition strategies allow weighting the models.
We implement and evaluate three weighting strategies, based on skill progress, demonstration variance, and parallel sampling.

\begin{figure*}
    \centering
    \begin{subfigure}[t]{0.24\textwidth}
        \centering
        \includegraphics[width=\textwidth]{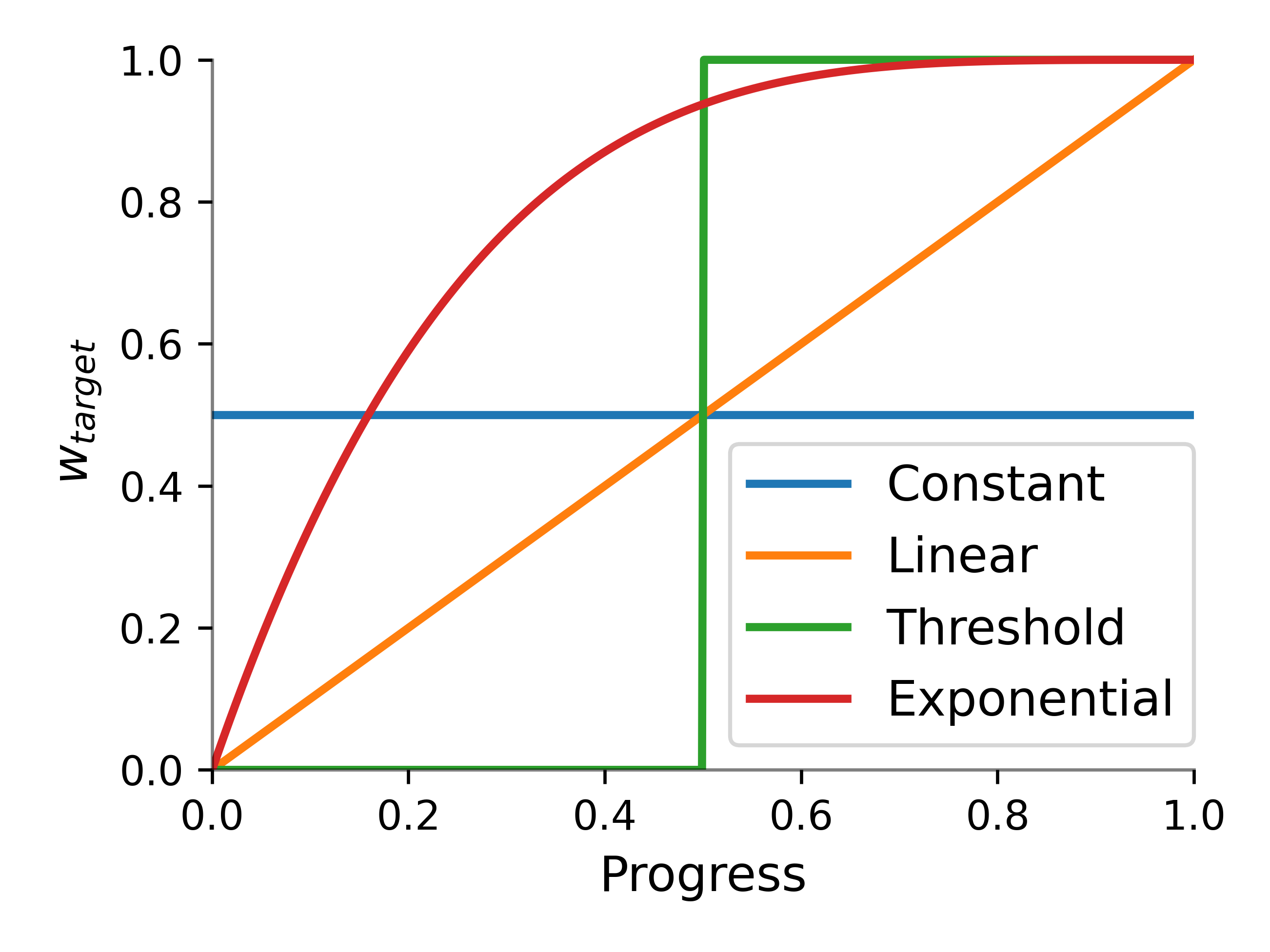}
        \caption{Progress-based weighting schedules}
        \label{fig:progress_weighting}
    \end{subfigure}
    \begin{subfigure}[t]{0.24\textwidth}
        \centering
        \includegraphics[width=\textwidth]{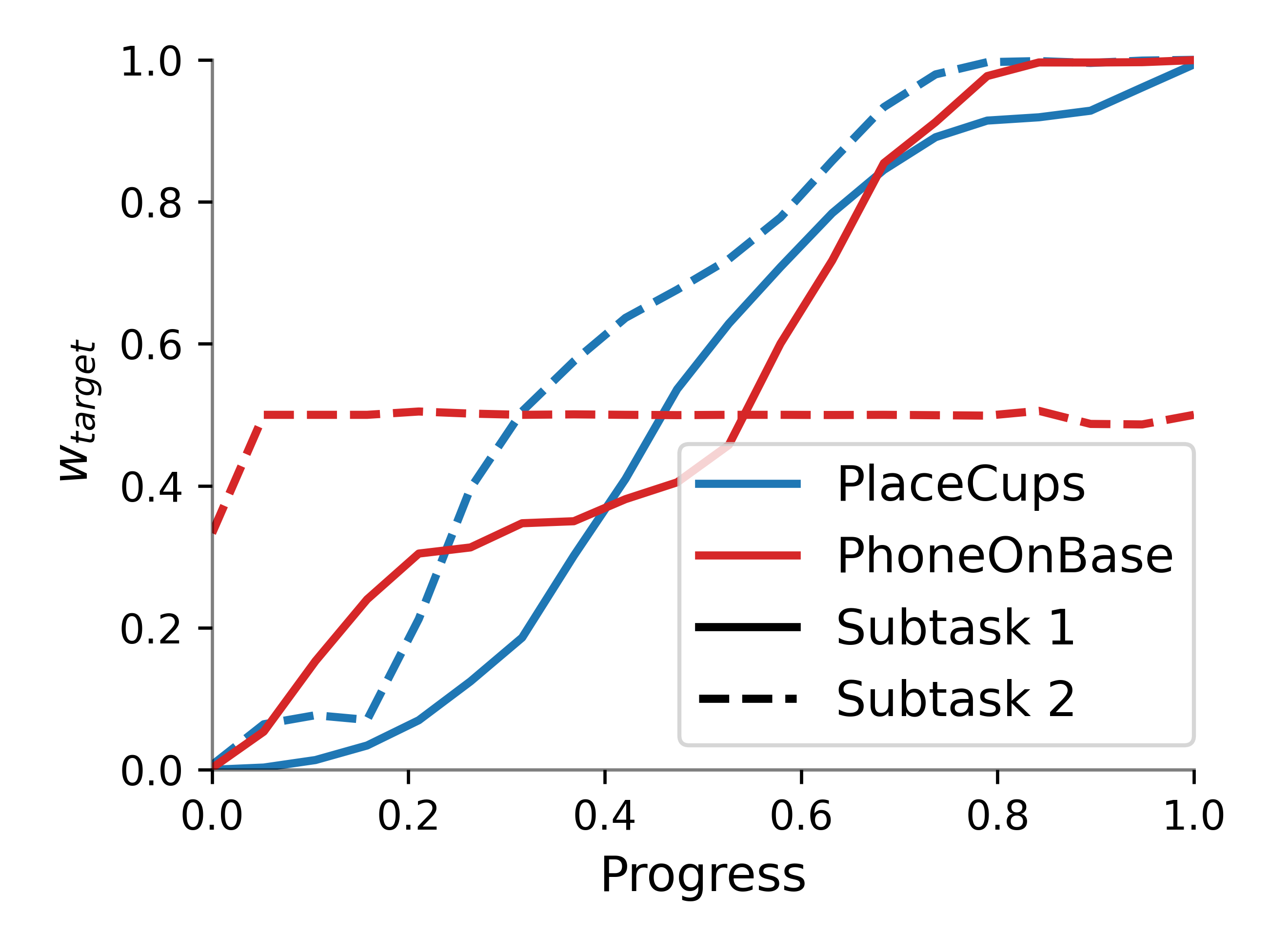}
        \caption{Variance-based schedules (data)}
        \label{fig:variance_weighting_train}
    \end{subfigure}
    \begin{subfigure}[t]{0.24\textwidth}
        \centering
        \includegraphics[width=\textwidth]{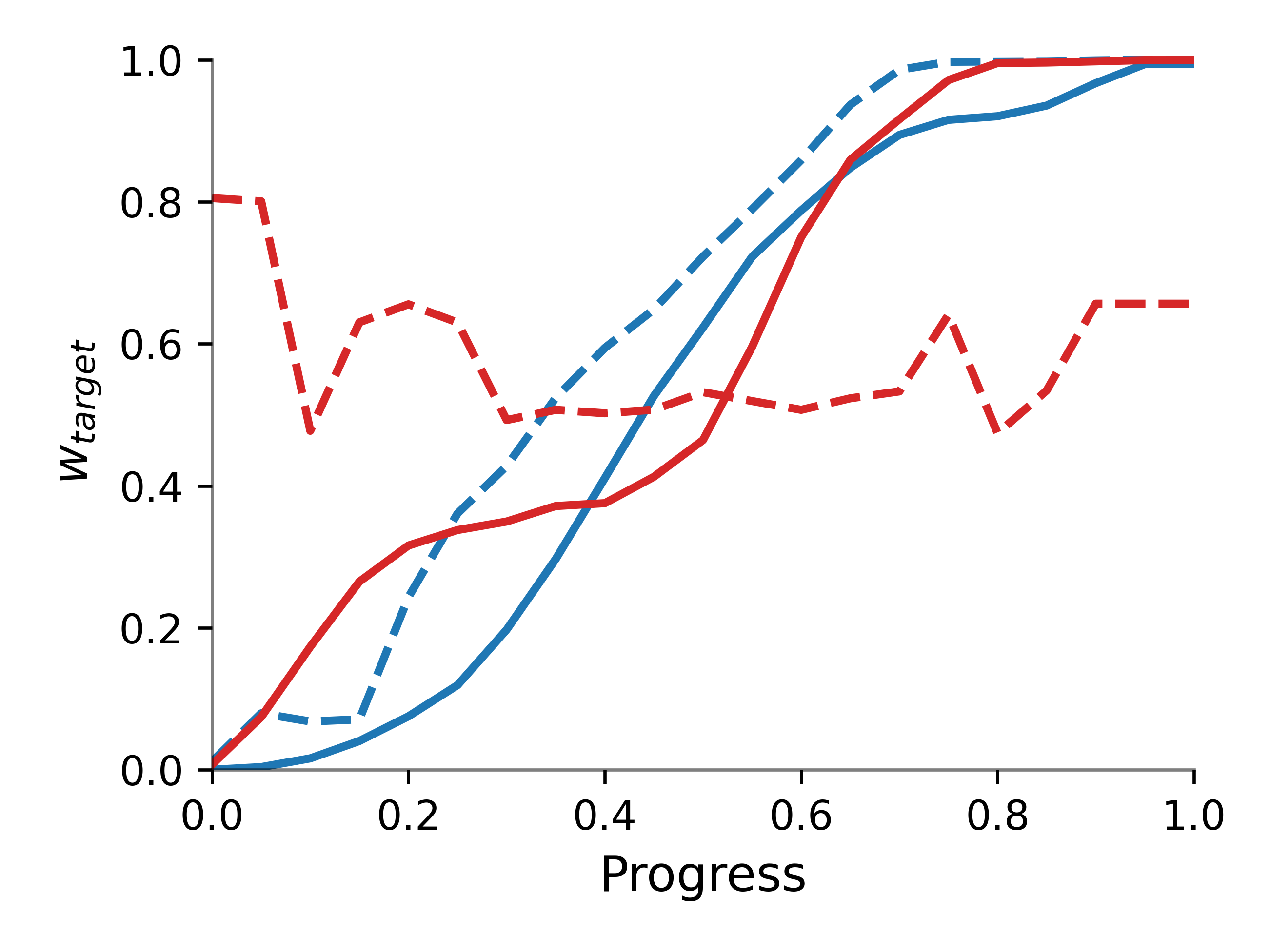}
        \caption{Variance-based schedules (inferred)}
        \label{fig:variance_weighting_infer}
    \end{subfigure}
    \begin{subfigure}[t]{0.24\textwidth}
        \centering
        \includegraphics[width=\textwidth]{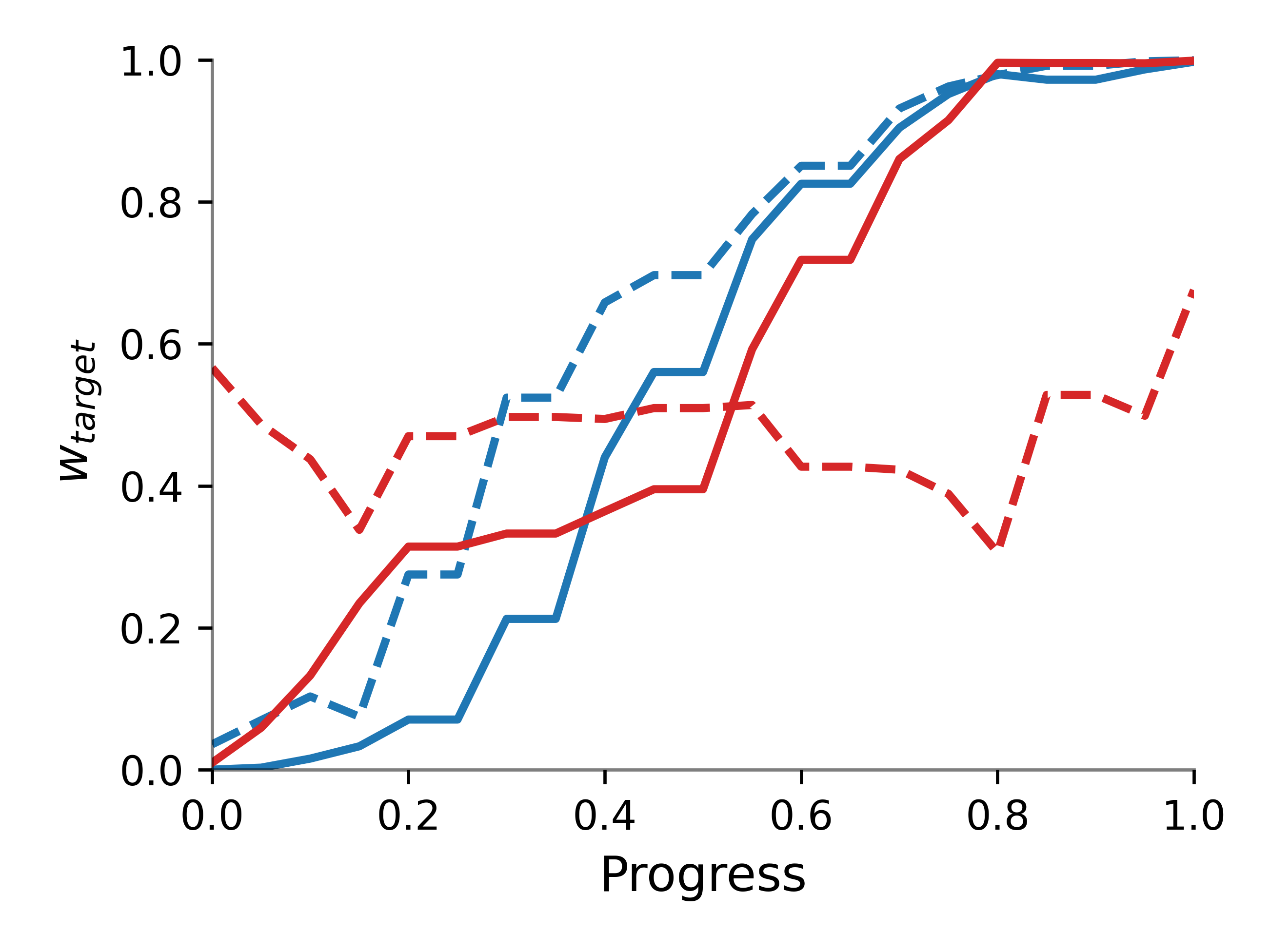}
        \caption{Particle-variance-based schedules}
        \label{fig:multi_particle_weighting}
    \end{subfigure}

    \caption{Stream weighting schedules.
    The schedules derived from the demonstration variance, shown in \subfiguresubref{fig:variance_weighting_train}, are reproduced mostly faithfully by the models' own logvar prediction, as shown in \subfiguresubref{fig:variance_weighting_infer}.
    The same holds for the variances estimates derived from parallel sampling, as shown in \subfiguresubref{fig:multi_particle_weighting}.
    }
    \label{fig:weighting_strategies}
    \vspace{-0.4cm}
\end{figure*}

{\parskip=2pt
\noindent\textit{Progress-Based Scheduling:} 
As introduced in \secref{sec:back_obj_centric}, our policies predict a progress value alongside the end-effector pose.
For two-stream policies, the following schedules map the skill progress \(p_t\) to the stream weights \(w_1, w_2\).
All schedules are shown in \figref{fig:progress_weighting}.
With \(w_2=1-w_1\), we define}
\begin{itemize}
    \item Constant: \(w_1=\frac{1}{2}\)
    \item Threshold: \(w_1 = \mathds{1}_{p_t<0.5}\)
    \item Linear: \(w_1 = p_t\)
    \item Exponential: \(w_1 = (1 - p_t)^4\).
\end{itemize}
While serving as simple baselines, these schedules are limited to two streams.
\rebuttal{They further require knowledge on the order of the streams, i.e.\ which object is manipulated when.
While the manipulation order can be estimated~\cite{heppert2024ditto}, }
a more generalized approach estimates the weights in a data-driven manner instead.

{\parskip=2pt
\noindent\textit{Variance-Based Weighting:} 
Consider the Gaussian policies shown in \figref{fig:multi_frame}.
In a product-of-Gaussians model, each component is weighted by its variance.
Single-particle flow matching does not provide variance estimates.
To address this, we train each local model \(f\) to predict its own log variance \(\boldsymbol\psi_{\pose_\ee^{(f)}}\), using supervision from a discrete-time Gaussian process~\cite{von2025unreasonable}.
In other words, the local models learn to estimate their own uncertainty.}\footnote{
We assume unimodality of the object-centric data distribution here.
This approach can be extended to multimodal scenarios by using a Mixture of Discrete-time Gaussian Processes for supervision of the variance estimate~\cite{von2025unreasonable}.
As \figref{fig:toy_comp} illustrates, flow composition encourages convergence of multiple streams to a common mode.
We forego this extension here as the parallel sampling approach, we introduce next, natively applies to multimodal cases.
}
Note that we do not predict the full covariance matrix of the pose, but only its main diagonal.
We evaluate two variants: predicting all six diagonal elements independently (6D), and a simplified 2D version that groups variance for position and orientation separately.

During inference, we transform the predicted log variances \(\{\boldsymbol\psi_{\pose_\ee^{(f)}}\}_{f=1}^F\) into world frame, yielding \(\{\boldsymbol\psi_{\pose_{\ee,f}}\}_{f=1}^F\) and assign
\begin{equation}\label{eq:var_weight}
    \boldsymbol w_f \propto\exp\left(-\boldsymbol\psi_{\pose_{\ee,f}}\right).
\end{equation}
This flexible, data-driven approach affords an arbitrary number of streams and does not require manual weight engineering.
In contrast to the schedules, \(\boldsymbol w_f\) is now a weight vector, enabling a weighting per data dimension.
Finally, as inference is still single-particle, this approach increases computational cost only marginally during both training and inference.

{\parskip=2pt
\noindent\textit{Parallel Sampling:} 
Instead of training the local models to predict their own uncertainties, we can estimate uncertainty at inference time.
At the price of increased inference cost, we draw multiple samples \(\{{\hat\flowVariable}_{0, k}\}_{k=1}^K\sim p_0\) and integrate them in parallel.
From the resulting set of particles \(\{{\hat\flowVariable}_{1, k}^{(f)}\}_{k=1}^K\), we estimate each stream’s variance \(\boldsymbol\sigma^2_{\pose_\ee^{(f)}}\) and assign weights inversely proportional, i.e.,}
\begin{equation}\label{eq:part_weight}
    \boldsymbol w_f^{-1} \propto\boldsymbol\sigma^2_{\pose_\ee^{(f)}}.
\end{equation}

A full robot trajectory \((\pose_s)_{s=1}^S\) typically requires repeated policy predictions in a receding horizon manner.
The flow models \(\boldsymbol v_f\) are conditioned on the current end-effector pose \(\pose_\ee^{(f)}\).
As a result, particle diversity collapses as the models observe the same deterministic input for each particle.
To mitigate this, we condition each \(\boldsymbol v_f\) on a virtual end-effector pose \({\smash{\hat\pose}}_{\ee, k}^{(f)}\), specific to each particle \(\hat\flowVariable_{t,k}^{(f)}\), instead of the true pose.
For the first prediction in a robot trajectory, we initialize \(\{\smash{\hat{\pose}}_{\ee, k}^{(f)}\}_{k=1}^K\) by sampling initial end-effector poses from the training set.
For all subsequent predictions, we reuse the previously integrated particle state \(\{{\hat\flowVariable}_{1, k}^{(f)}\}_{k=1}^K\) as virtual poses, 
i.e., we retain the particle population across the entire \emph{robot} trajectory.
This strategy preserves particle diversity over time, allowing us to continuously estimate each model’s uncertainty throughout the robot trajectory.

\section{Experiments}\label{sec:exp}
We evaluate \ourmethod's efficacy using a suite of simulated and real-world policy learning experiments.
First, we perform extensive experiments in RLBench~\cite{james2019rlbench} to establish statistically significant results across a diverse range of tasks.
Subsequently, we investigate the scaling properties of \ourmethod.
We then perform comprehensive ablation studies, evaluating the impact of our two different composition strategies, eight weighting strategies, and other key design choices.
Finally, we validate our results on a real Franka Emika Panda robot.

\subsection{Policy Learning in Simulation}\label{sec:rlbench}
We evaluate MSG using four single-object tasks, as well as \rebuttal{five} multi-object tasks, shown in \figref{fig:tasks}.
The single-object tasks all involve articulated objects, hence requiring very precise actions, lest the robot slip off, for example, the handle of the microwave.
The multi-object tasks, in contrast, involve longer horizons and require the composition of more local models.
Some of them, such as \texttt{InsertOntoSquarePeg}, involve both multiple objects \emph{and} high precision requirements.
We automatically~\cite{vonhartz2024art} segment all tasks into two skills, which are sequenced as described in \secref{sec:back_obj_centric}.
For the single-object tasks, the first skill is approaching the object, whereas the second skill is interacting with the object (e.g., opening the door).
For the multi-object tasks, the first skill includes the first object (e.g., picking), whereas the second skill involves both (e.g.,\ placing).
\rebuttal{The objects involved determine which streams are used per skill.}
All policies are trained using standard hyperparameters~\cite{chi2023diffusionpolicy, chisari2024flowmatch, rana2024affordancecentricpolicylearningsample}.
\tabref{tab:success_rates_rlbench} compares the policy success of MSG to a series of baselines, including naive Flow Matching~\cite{lipman2022flow} conditioned on the object poses, object-centric learning in start or goal frame, and ACPL (with oriented object frame)~\cite{rana2024affordancecentricpolicylearningsample}.
MSG's reported success rates are for the exponential schedule.
Success rates for all other weighting strategies are reported in \secref{sec:ablation}.

\begin{table*}
        \caption{Policy success rates (mean and standard deviation) on RLBench using 10 demonstrations. }\label{tab:success_rates_rlbench}
        \centering
        \setlength{\tabcolsep}{3.25pt}
        \begin{threeparttable}
        \begin{tabular}{l ccccc cccccc}
            \toprule
            \multirow{2}{*}{\textbf{Method}} & \multicolumn{5}{c}{\textbf{Single Object Tasks}} & \multicolumn{5}{c}{\textbf{Multi Object Tasks}} \\
            \cmidrule(lr){2-6} \cmidrule(lr){7-12}
            & \makecell{Open \\ Drawer} & \makecell{Open \\ Microwave} & \makecell{Toilet \\ SeatUp} & \makecell{Turn \\ Tap} & Avg. & \makecell{Stack \\ Wine} & \makecell{Place \\ Cups} & \makecell{Phone \\ OnBase} & \makecell{InsertOnto \\ SquarePeg} &\makecell{\rebuttal{Stack} \\ \rebuttal{Blocks}} & Avg. \\
            \midrule
            Global Frame~\cite{lipman2022flow, chi2023diffusionpolicy} & 0.65$\pm$0.03 & 0.30$\pm$0.06 & 0.88$\pm$0.02 & 0.43$\pm$0.03 & 0.57 & 0.63$\pm$0.02 & 0.06$\pm$0.02 & 0.27$\pm$0.03 & 0.00$\pm$0.00 & \rebuttal{0.00$\pm$0.00} & 0.19\\
            End-Effector Frame & 0.62$\pm$0.06 & 0.23$\pm$0.05 & 0.81$\pm$0.01 & 0.43$\pm$0.04 & 0.52 & 0.87$\pm$0.01 & 0.06$\pm$0.02 & 0.29$\pm$0.03 & 0.01$\pm$0.00 & \rebuttal{0.00$\pm$0.00} & 0.25\\
            Object Frame & 0.67$\pm$0.07 & 0.74$\pm$0.00 & \textbf{0.94$\pm$0.01} & 0.49$\pm$0.04 & 0.71 & 0.86$\pm$0.00 & 0.35$\pm$0.01 & 0.36$\pm$0.04 & 0.24$\pm$0.02 & \rebuttal{0.04$\pm$0.01} & 0.37\\
            ACPL (Oriented Frame)~\cite{rana2024affordancecentricpolicylearningsample} & 0.63$\pm$0.07 & 0.65$\pm$0.09 & 0.90$\pm$0.01 & 0.53$\pm$0.01 & 0.68 & 0.85$\pm$0.04 & 0.33$\pm$0.00 & 0.45$\pm$0.02 & 0.09$\pm$0.02 & \rebuttal{0.01$\pm$0.01} & 0.35\\
            \cmidrule{1-12}
            \textbf{\ourmethod Ensemble (Ours)} & 0.77$\pm$0.07 & 0.83$\pm$0.00 & 0.92$\pm$0.02 & 0.66$\pm$0.08 & 0.80 & 0.97$\pm$0.00 & 0.43$\pm$0.04 & 0.44$\pm$0.01 & 0.39$\pm$0.02 & \rebuttal{0.30$\pm$0.02} & 0.51\\
            \textbf{\ourmethod Flow Comp.\ (Ours)} & 0.66$\pm$0.04 & 0.84$\pm$0.02 & \textbf{0.94$\pm$0.02} & \textbf{0.70$\pm$0.08} & 0.79 & 0.95$\pm$0.01 & 0.43$\pm$0.01 & 0.45$\pm$0.02 & 0.45$\pm$0.02 & \rebuttal{0.31$\pm$0.01} & 0.52 \\
            \textbf{\ourmethod Flow MCMC (Ours)} & \textbf{0.97$\pm$0.01} & \textbf{0.95$\pm$0.03} & 0.91$\pm$0.02 & 0.69$\pm$0.03 & \textbf{0.88} & \textbf{0.98$\pm$0.01} & \textbf{0.57$\pm$0.04} & \textbf{0.52$\pm$0.01} & \textbf{0.52$\pm$0.02} & \rebuttal{\textbf{0.33$\pm$0.03}} & \textbf{0.58}\\
            \bottomrule
        \end{tabular}
        \begin{tablenotes}[para,flushleft]
           \footnotesize  
           Bold values indicate the best-performing model for each task.
           All results are averaged over three runs.
           MSG values are reported for the exponential schedule.
         \end{tablenotes}
        \end{threeparttable}
\end{table*}

All three variants of MSG consistently outperform all baseline methods across all tasks.
On average, compared to standard Flow Matching, MSG improves policy success by 31 percentage points on the single-object tasks, and by 41 percentage points on the multi-object tasks.
In relative terms and averaged over both, that is an improvement of 89\%.
In contrast, ACPL, with its oriented object frame, performs slightly worse than the object-centric baseline.
By improving generalization, multi-stream learning boosts performance more on tasks with larger variance in object poses.
\texttt{ToiletSeatUp}, for instance, has very little variance, leading to little advantage over single-stream learning.
In contrast, \texttt{TurnTap}, \texttt{InsertOntoSquarePeg} have large variance (and require high precision), leading to large performance gains of MSG.

Our \ourmethod Ensemble performs exceptionally strong across the board.
Flow Composition only outperforms it when coupled with MCMC.
MCMC mostly helps on tasks with high precision requirements, like \texttt{OpenDrawer}, \texttt{OpenMicrowave}, \texttt{PlaceCups}, and \texttt{InsertOntoSquarePeg}.
Because the ensemble-based composition is significantly easier to set up and does not interfere with other techniques, such as data normalization, we recommend it as the first method to try for unimodal tasks.
\rebuttal{Moreover, the ensemble increases inference time by only 12\% over single-frame policies, while flow composition adds 69\%.
MCMC adds an additional four-fold overhead.
Full inference times are listed in \secref{sec:inf_cost}.}

\begin{figure}
    \centering
    \begin{subfigure}[t]{0.49\linewidth}
        \centering
        \includegraphics[width=\linewidth]{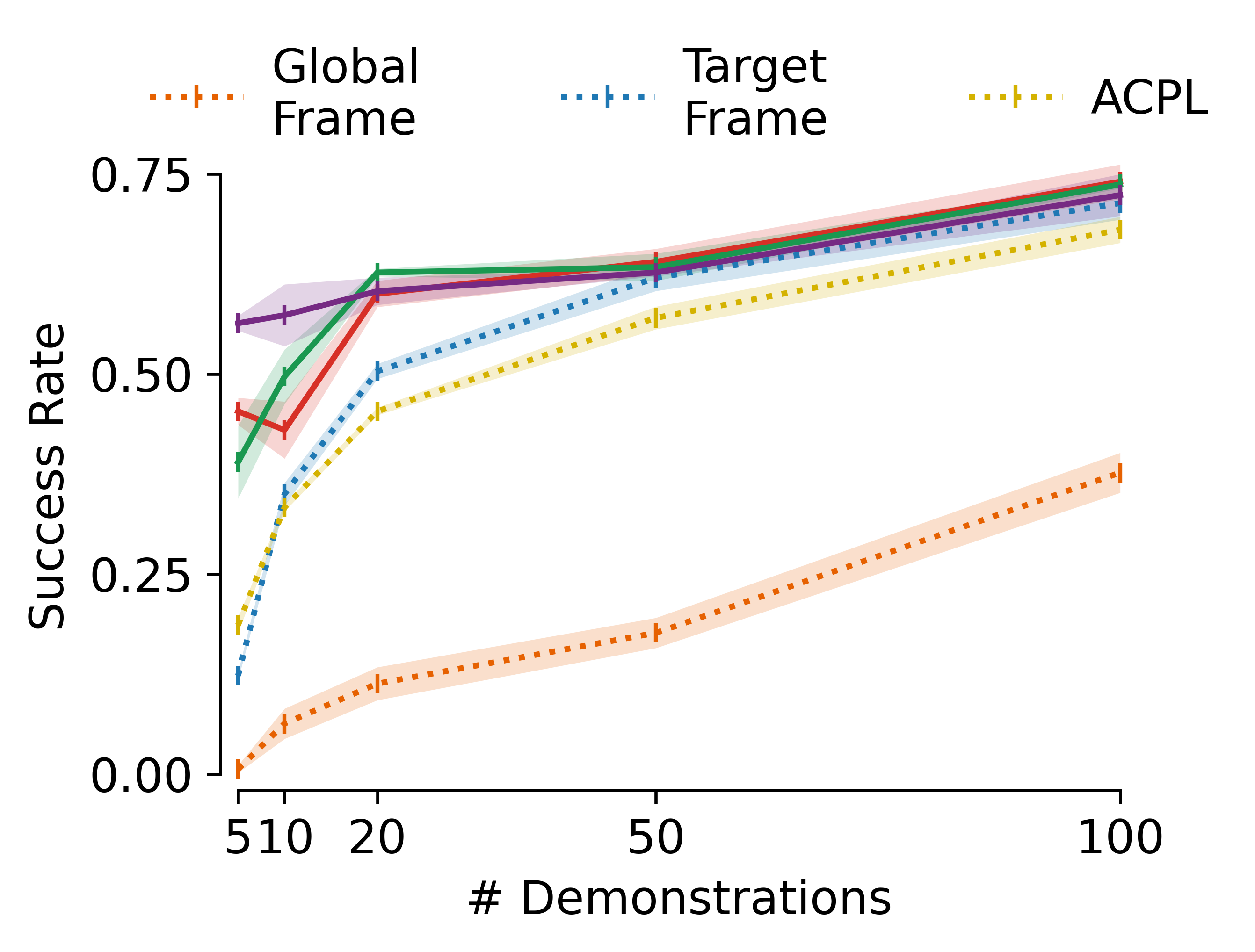}
        \subcaption{PlaceCups}
        \label{fig:data_scaling_placecups}
    \end{subfigure}%
    \hfill
    \begin{subfigure}[t]{0.49\linewidth}
        \centering
        \includegraphics[width=\linewidth]{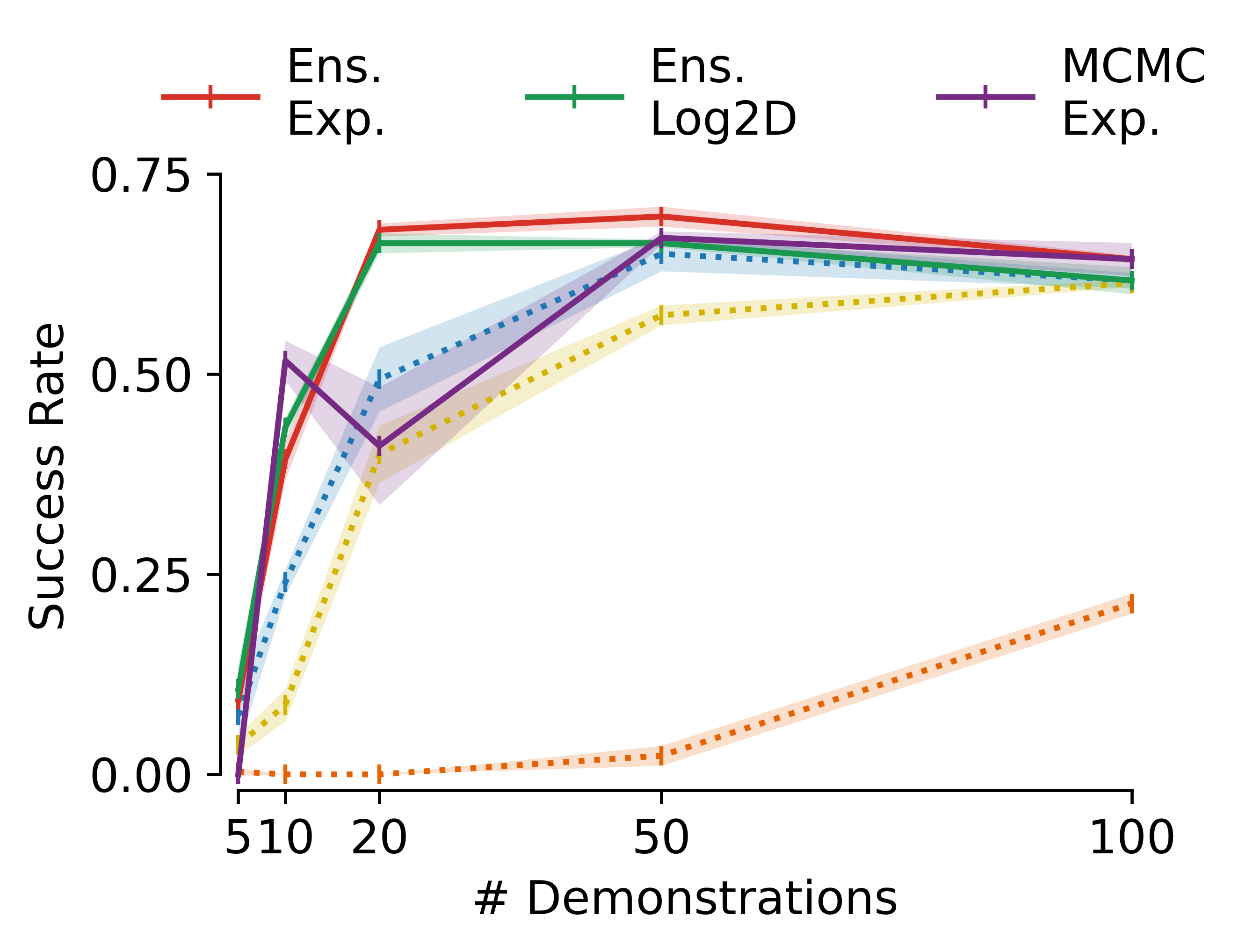}
        \subcaption{InsertOntoSquarePeg}
        \label{fig:data_scaling_insertontosquarepeg}
    \end{subfigure}
    \caption{
    Data scaling analysis. \ourmethod excels even when more data is available.
    }
    \label{fig:data_scaling}
    \vspace{-6mm}
\end{figure}

\subsection{Data Scaling}
\figref{fig:data_scaling} shows how the evaluated methods scale with increased dataset size.
Two observations are noteworthy.
First, MSG learns strong policies even from only five demonstrations.
With as little as five demonstrations, MSG outperforms the standard Flow Matching policy trained on 100 demonstrations.
Second, while the performance gap shrinks with an increased number of demonstrations, it does not vanish.
MSG is also effective in large data regimes.
All models decreasing in performance on \texttt{InsertOntoSquarePeg} beyond 50 demonstrations are due to overfitting.
We train using a fixed number of episodes to ensure a fair evaluation. 
Fewer episodes remedy overfitting.

\subsection{Weighting Strategies and Ablation Study}\label{sec:ablation}

\begin{table*}
        \caption{Comparison of stream weighting strategies on RLBench tasks. }\label{tab:schedules_rlbench}
        \centering
        \setlength{\tabcolsep}{3.25pt}
        \begin{threeparttable}
        \begin{tabular}{ll ccccc cccccc}
            \toprule
            \multirow{2}{*}{\textbf{Composition}} & \multirow{2}{*}{\textbf{Weighting}} & \multicolumn{5}{c}{\textbf{Single Object Tasks}} & \multicolumn{6}{c}{\textbf{Multi Object Tasks}} \\
            \cmidrule(lr){3-7} \cmidrule(lr){8-13}
            & & \makecell{Open \\ Drawer} & \makecell{Open \\ Microwave} & \makecell{Toilet \\ SeatUp} & \makecell{Turn \\ Tap} & Avg. & \makecell{Stack \\ Wine} & \makecell{Place \\ Cups} & \makecell{Phone \\ OnBase} & \makecell{InsertOnto \\ SquarePeg} & \makecell{Stack \\ Blocks} & Avg. \\
            \midrule
            Ensemble & Constant & 0.71$\pm$0.03 & 0.63$\pm$0.05 & 0.87$\pm$0.02 & 0.51$\pm$0.04 & 0.68 & 0.92$\pm$0.00 & 0.15$\pm$0.00 & 0.36$\pm$0.02 & 0.01$\pm$0.00 & \rebuttal{0.00$\pm$0.00} & \rebuttal{0.29}\\
            & Threshold         & 0.59$\pm$0.02 & 0.81$\pm$0.02 & 0.91$\pm$0.01 & 0.61$\pm$0.05 & 0.73 & 0.96$\pm$0.01 & 0.46$\pm$0.04 & 0.44$\pm$0.02 & 0.37$\pm$0.02 & \rebuttal{0.26$\pm$0.02} & \rebuttal{0.50}\\
            & Linear            & 0.62$\pm$0.03 & 0.81$\pm$0.04 & 0.90$\pm$0.02 & 0.62$\pm$0.06 & 0.73 & 0.96$\pm$0.01 & 0.30$\pm$0.04 & 0.46$\pm$0.03 & 0.17$\pm$0.04 & \rebuttal{0.09$\pm$0.00} & \rebuttal{0.39}\\
            & Exponential       & 0.77$\pm$0.07 & 0.83$\pm$0.00 & 0.92$\pm$0.02 & 0.66$\pm$0.08 & 0.80 & 0.97$\pm$0.00 & 0.43$\pm$0.04 & 0.44$\pm$0.01 & 0.39$\pm$0.02 & \rebuttal{0.30$\pm$0.02} & \rebuttal{0.51}\\
            & Logvar (6D)       & 0.59$\pm$0.06 & 0.74$\pm$0.04 & 0.89$\pm$0.02 & 0.60$\pm$0.06 & 0.70 & 0.95$\pm$0.00 & 0.38$\pm$0.04 & 0.34$\pm$0.03 & 0.23$\pm$0.07 & \rebuttal{0.07$\pm$0.01} & \rebuttal{0.39}\\
            & Logvar (2D)       & 0.71$\pm$0.06 & 0.86$\pm$0.05 & 0.90$\pm$0.02 & 0.66$\pm$0.04 & 0.78 & 0.97$\pm$0.00 & 0.50$\pm$0.03 & 0.38$\pm$0.04 & {0.43$\pm$0.01} & \rebuttal{0.26$\pm$0.02} & \rebuttal{0.51}\\
            & Particle (6D)     & 0.70$\pm$0.10 & 0.85$\pm$0.07 & 0.89$\pm$0.02 & 0.59$\pm$0.10 & 0.76 & 0.97$\pm$0.01 & 0.30$\pm$0.02 & 0.41$\pm$0.03 & 0.27$\pm$0.02 & \rebuttal{0.05$\pm$0.00} & \rebuttal{0.40}\\
            & Particle (2D)     & 0.66$\pm$0.07 & 0.80$\pm$0.05 & 0.90$\pm$0.00 & 0.63$\pm$0.09 & 0.75 & 0.97$\pm$0.01 & 0.44$\pm$0.07 & 0.42$\pm$0.01 & 0.41$\pm$0.00 & \rebuttal{0.19$\pm$0.01} & \rebuttal{0.49}\\
            \cmidrule{1-13}
            Flow 
            & Constant          & 0.68$\pm$0.04 & 0.64$\pm$0.00 & 0.89$\pm$0.01 & 0.53$\pm$0.03 & 0.68 & 0.88$\pm$0.04 & 0.15$\pm$0.03 & 0.38$\pm$0.01 & 0.03$\pm$0.01 & \rebuttal{0.00$\pm$0.00} & \rebuttal{0.29}\\
            & Threshold         & 0.63$\pm$0.05 & 0.80$\pm$0.03 & 0.88$\pm$0.02 & 0.66$\pm$0.06 & 0.74 & 0.94$\pm$0.01 & 0.45$\pm$0.04 & 0.42$\pm$0.01 & 0.39$\pm$0.03 & \rebuttal{0.24$\pm$0.02} & \rebuttal{0.49}\\
            & Linear            & 0.65$\pm$0.03 & 0.81$\pm$0.02 & 0.89$\pm$0.01 & 0.67$\pm$0.05 & 0.76 & 0.97$\pm$0.01 & 0.33$\pm$0.02 & 0.47$\pm$0.02 & 0.17$\pm$0.04 & \rebuttal{0.12$\pm$0.02} & \rebuttal{0.41}\\
            & Exponential       & 0.66$\pm$0.04 & 0.84$\pm$0.02 & \textbf{0.94$\pm$0.02} & 0.70$\pm$0.08 & 0.79 & 0.95$\pm$0.01 & 0.43$\pm$0.01 & 0.45$\pm$0.02 & 0.45$\pm$0.02 & \rebuttal{0.31$\pm$0.01} & \rebuttal{0.52}\\
            & Logvar (6D)       & 0.61$\pm$0.01 & 0.75$\pm$0.04 & 0.89$\pm$0.03 & 0.72$\pm$0.04 & 0.74 & 0.95$\pm$0.01 & 0.42$\pm$0.04 & 0.40$\pm$0.02 & 0.23$\pm$0.02 & \rebuttal{0.06$\pm$0.03} & \rebuttal{0.41}\\
            & Logvar (2D)       & 0.65$\pm$0.07 & 0.81$\pm$0.04 & 0.91$\pm$0.02 & 0.71$\pm$0.04 & 0.77 & 0.94$\pm$0.01 & 0.47$\pm$0.08 & 0.44$\pm$0.02 & 0.39$\pm$0.03 & \rebuttal{0.30$\pm$0.03} & \rebuttal{0.51}\\
            & Particle (6D)     & \rebuttal{0.72$\pm$0.05} & \rebuttal{0.82$\pm$0.04} & \rebuttal{0.88$\pm$0.00} & \rebuttal{0.66$\pm$0.05} & \rebuttal{0.77} & \rebuttal{0.95$\pm$0.01} & \rebuttal{0.39$\pm$0.05} & \rebuttal{0.47$\pm$0.01} & \rebuttal{0.25$\pm$0.04} & \rebuttal{0.06$\pm$0.02} & \rebuttal{0.42}\\
            & Particle (2D)     & \rebuttal{0.69$\pm$0.07} & \rebuttal{0.85$\pm$0.01} & \rebuttal{0.91$\pm$0.00} & \rebuttal{0.66$\pm$0.09} & \rebuttal{0.78} & \rebuttal{0.97$\pm$0.00} & \rebuttal{0.48$\pm$0.02} & \rebuttal{0.45$\pm$0.02} & \rebuttal{0.40$\pm$0.04} & \rebuttal{0.20$\pm$0.03} & \rebuttal{0.50}\\
            \cmidrule{1-13}
            Flow MCMC  %
            & Constant          & 0.92$\pm$0.01 & 0.78$\pm$0.04 & 0.86$\pm$0.01 & 0.46$\pm$0.09 & 0.76 & 0.93$\pm$0.00 & 0.17$\pm$0.02 & 0.45$\pm$0.02 & 0.01$\pm$0.01 & \rebuttal{0.00$\pm$0.00} & \rebuttal{0.31}\\
            & Threshold         & 0.91$\pm$0.02 & 0.92$\pm$0.02 & 0.87$\pm$0.03 & 0.67$\pm$0.03 & 0.84 & 0.97$\pm$0.01 & 0.54$\pm$0.04 & {0.52$\pm$0.01} & 0.46$\pm$0.03 & \rebuttal{0.29$\pm$0.02} & \rebuttal{0.55}\\
            & Linear            & 0.92$\pm$0.01 & 0.91$\pm$0.03 & 0.88$\pm$0.00 & 0.67$\pm$0.03 & 0.85 & 0.97$\pm$0.02 & 0.44$\pm$0.08 & 0.51$\pm$0.00 & 0.21$\pm$0.01 & \rebuttal{0.16$\pm$0.01} & \rebuttal{0.46}\\
            & Exponential       & \textbf{0.97$\pm$0.01} & 0.95$\pm$0.03 & 0.91$\pm$0.02 & 0.69$\pm$0.03 & 0.88 & \textbf{0.98$\pm$0.01} & \textbf{0.57$\pm$0.04} & {0.52$\pm$0.01} & \textbf{0.52$\pm$0.02} & \rebuttal{0.33$\pm$0.03} & \rebuttal{\textbf{0.58}} \\
            & Logvar (6D)       & 0.89$\pm$0.10 & 0.92$\pm$0.05 & 0.88$\pm$0.01 & 0.72$\pm$0.06 & 0.85 & 0.96$\pm$0.03 & 0.56$\pm$0.05 & 0.42$\pm$0.03 & 0.14$\pm$0.04 & \rebuttal{0.08$\pm$0.01} & \rebuttal{0.43}\\
            & Logvar (2D)       & 0.95$\pm$0.02 & \textbf{0.97$\pm$0.00} & 0.90$\pm$0.01 & \textbf{0.73$\pm$0.01} & \textbf{0.89} & 0.97$\pm$0.00 & \textbf{0.57$\pm$0.02} & 0.47$\pm$0.00 & 0.38$\pm$0.02 & \rebuttal{\textbf{0.39$\pm$0.02}} & \rebuttal{0.56} \\
            & Particle (6D)     & \rebuttal{\textbf{0.97$\pm$0.01}} & \rebuttal{0.96$\pm$0.01} & \rebuttal{0.90$\pm$0.00} & \rebuttal{0.67$\pm$0.07} & \rebuttal{0.88} & \rebuttal{0.97$\pm$0.03} & \rebuttal{0.53$\pm$0.04} & \rebuttal{\textbf{0.53$\pm$0.01}} & \rebuttal{0.31$\pm$0.04} & \rebuttal{0.06$\pm$0.01} & \rebuttal{0.48}\\
            & Particle (2D)     & \rebuttal{0.92$\pm$0.00} & \rebuttal{0.89$\pm$0.04} & \rebuttal{0.89$\pm$0.02} & \rebuttal{0.65$\pm$0.07} & \rebuttal{0.84} & \rebuttal{0.96$\pm$0.00} & \rebuttal{\textbf{0.57$\pm$0.00}} & \rebuttal{0.51$\pm$0.00} & \rebuttal{0.38$\pm$0.02} & \rebuttal{0.31$\pm$0.00} & \rebuttal{0.55}\\
            \bottomrule
        \end{tabular}
        \begin{tablenotes}[para,flushleft]
            \footnotesize  
            All policies are end-effector conditioned. Flow Composition uses local noise, whereas MCMC uses its custom prior.
            These choices are ablated in \tabref{tab:ablation_rlbench}.
          \end{tablenotes}
        \end{threeparttable}
        \vspace{-0.3cm}
\end{table*}

\begin{table}
        \caption{\ourmethod ablation study on RLBench tasks: policy success rates.}\label{tab:ablation_rlbench}
        \centering
        \setlength{\tabcolsep}{4pt}
        \begin{threeparttable}
        \begin{tabular}{l cccccc}
            \toprule
            \multirow{2}{*}{\textbf{Variant}} & \multicolumn{4}{c}{\textbf{Schedule}} & \multicolumn{2}{c}{\textbf{Logvar}} \\
            \cmidrule(lr){2-5}\cmidrule(lr){6-7}
            & Con. & Thre. & Lin. & Exp. & 6D & 2D \\

            \midrule
            \textbf{Flow Composition}                & \rebuttal{0.46} & \rebuttal{0.60} & \rebuttal{0.56} & \rebuttal{0.64} & \rebuttal{0.56} & \rebuttal{0.62} \\
            └─ w/o\, Custom Prior                    & \rebuttal{0.39} & \rebuttal{0.53} & \rebuttal{0.48} & \rebuttal{0.54} & \rebuttal{0.47} & \rebuttal{0.54} \\
            └─ w/o\, Sample Matching                 & \rebuttal{0.04} & \rebuttal{0.60} & \rebuttal{0.16} & \rebuttal{0.33} & \rebuttal{0.17} & \rebuttal{0.16} \\
            └─ w/o\, Conditioning                    & \rebuttal{0.33} & \rebuttal{0.55} & \rebuttal{0.51} & \rebuttal{0.63} & \rebuttal{0.46} & \rebuttal{0.55} \\
            └─ with MCMC-Matched Steps               & \rebuttal{0.48} & \rebuttal{0.62} & \rebuttal{0.59} & \rebuttal{0.64} & \rebuttal{0.56} & \rebuttal{0.63} \\
            \textbf{Flow Composition\ MCMC}          & \rebuttal{0.51} & \rebuttal{0.68} & \rebuttal{0.63} & \rebuttal{0.71} & \rebuttal{0.62} & \rebuttal{0.70} \\
            └─ with Mixture Prior                    & \rebuttal{0.39} & \rebuttal{0.49} & \rebuttal{0.47} & \rebuttal{0.51} & \rebuttal{0.43} & \rebuttal{0.54} \\
            └─ w/o\, Custom Prior                    & \rebuttal{0.47} & \rebuttal{0.63} & \rebuttal{0.60} & \rebuttal{0.65} & \rebuttal{0.60} & \rebuttal{0.68} \\
            └─ w/o\, Conditioning                    & \rebuttal{0.46} & \rebuttal{0.63} & \rebuttal{0.58} & \rebuttal{0.70} & \rebuttal{0.53} & \rebuttal{0.63} \\
            \bottomrule
        \end{tabular}
        \begin{tablenotes}[para,flushleft]
           \footnotesize  
           Results are averaged over all \rebuttal{nine} tasks and three runs per task. Standard deviations are omitted for brevity, but are around \(0.01\) for all models.
         \end{tablenotes}
    \end{threeparttable}
    \vspace{-0.6cm}
\end{table}

We report exhaustive results for all weighting strategies in \tabref{tab:schedules_rlbench}.
Across the three composition strategies, the exponential schedule tends to perform best.
\figref{fig:progress_weighting} explains why;
towards the end of the trajectory, the exponential schedule weights the target frame more strongly than, say, the linear schedule.
For tasks, such as \texttt{OpenDrawer}, this ensures precise grasping of the handle.
While such rigid scheduling works well for many tasks, it makes strong assumptions, which might not always hold.
For \texttt{PhoneOnBase}, for example, phone and base are always placed in the same relative pose to each other.
Therefore, in the second subtask of placing the phone on the base, the two local models are equally informative.
This can be seen in the variance weights extracted from the training data, shown in \figref{fig:variance_weighting_train}.
While \texttt{PlaceCups} is well captured by the exponential schedule, the weights for the second subtask of \texttt{PhoneOnBase} is close to being constant.
As \figref{fig:variance_weighting_infer} illustrates, the models trained to predict this variance information correctly reproduce it for the most part.
Consequently, the Logvar models achieve comparable performance to the progress-scheduled models, with the 2D (grouped) variant having a small but consistent advantage over the full 6D variant.
Finally, the weights derived from the particle variance are plotted in \figref{fig:multi_particle_weighting}.
They show the same patterns as the Logvar policies and achieve similar policy performance.
Hence, both logvar prediction and parallel sampling pose viable and more flexible alternatives to progress-based scheduling.
\rebuttal{High task variance favors the exponential schedule on \texttt{InsertOntoSquarePeg}, though this gap closes with additional demonstrations.
A full analysis is presented in \secref{sec:sched}.}

We also ablate key design choices in \tabref{tab:ablation_rlbench}.
As described in \secref{sec:model_comp}, matching the samples across both models is vital for policy success, as is our custom prior.
In contrast, the mixture prior alleviates distribution shift, but does not encourage converging flows.
Hence, it boosts policy performance for the flow composition, but worsens performance under MCMC.
Conditioning the flow models on the current end-effector pose (or the virtual pose of the particle) helps to produce consistent trajectory predictions, but its importance shrinks under MCMC.
\tabref{tab:schedules_rlbench} suggests that the benefit of MCMC is largely independent of the weighting strategy.
Even though the flow fields are not gradient-like, MCMC significantly boost policy success.
Moreover, \tabref{tab:ablation_rlbench} submits that this effect goes beyond the increased number of inference steps.

\subsection{Real World Experiments}

\begin{table}
    \begin{threeparttable}
        \caption{Real world policy success rates.}\label{tab:success_rates_real}
        \centering
        \setlength{\tabcolsep}{3.5pt}
        \begin{tabular}{l c c c c c c}
            \toprule
            \textbf{Method} & \makecell{Pick \\ AndPlace} &  \makecell{Pour \\ Drink} & \makecell{Sweep \\ Blocks} & \makecell{Open \\ Drawer} & \makecell{\rebuttal{StoreIn} \\ \rebuttal{Drawer}}& Avg.\\
            \midrule
            Global Frame & 0.04 & 0.00 & 0.24 & 0.52 & \rebuttal{0.00} & 0.16 \\
            Object Frame & 0.40 & 0.32 & 0.80 & 0.88 & \rebuttal{0.20} & 0.43 \\
            \textbf{\ourmethod Ensemble} & 0.72 & 0.64 & \textbf{0.88} & \textbf{0.96} & \rebuttal{0.64} & 0.77 \\
            \textbf{\ourmethod Flow MCMC} & \textbf{0.76} &  \textbf{0.84} & \textbf{0.88} &   \textbf{0.96} & \rebuttal{\textbf{0.68}} & \textbf{0.82} \\
            \bottomrule
        \end{tabular}
        \begin{tablenotes}[para,flushleft]
           \footnotesize  
           MSG values are reported for the exponential schedule.
         \end{tablenotes}
    \end{threeparttable}
   \vspace{-0.5cm}
\end{table}

We validate our approach with real-world experiments on a Franka Emika Panda robot using five representative tasks shown in \figref{fig:tasks}.
We compare both variants of \ourmethod (ensemble and flow) to the standard Flow Matching baseline, as well as to the best performing baseline, which is the single-stream object-centric model.
To estimate the object frames, we leverage a DINO-based keypoints predictor~\cite{vonhartz2024art, amir2021deep}.
\rebuttal{Alternative tracking methods are discussed in \secref{sec:kp_track}.}
Per task, we collect 10 demonstrations and evaluate each policy for 25 episodes.

\tabref{tab:success_rates_real} confirms our simulation results.
The standard generative policy fails to reliably solve any of the tasks given only 10 demonstrations.
The single-stream object-centric policy is more effective, but still exhibits limited generalization.
In contrast, both variants of \ourmethod achieve strong performance across the board, highlighting the generalization efficacy of multi-stream learning.
Flow composition performs best, but the ensemble does not lag far behind.
Qualitative examples of policy executions are shown in the supplementary video\rebuttal{, including a discussion of failure cases.}

Finally, estimating the object frames via DINO keypoints not only boosts policy success through object-centric learning but also enables zero-shot policy transfer to novel object instances and to novel and cluttered task environments.
This generalization has been studied quantitatively in prior work~\cite{amir2021deep, vonhartz2024art, von2025unreasonable}.
\rebuttal{We recreate the generalization study from~\cite{vonhartz2024art}, finding comparable results.
We provide qualitative examples in the supplementary video and quantitative results in \secref{sec:kp_track}.}

\section{Conclusion}
We presented \ourmethodfull (\ourmethod), a multi-stream, object-centric approach for generative policy learning.
\ourmethod learns robust policies from as few as five demonstrations, thus reducing data requirements and training duration by 95\% compared to standard generative policies.
\ourmethod is effective both when used as a straightforward ensemble and when performing more expressive flow composition.
For many tasks, \ourmethod can be used with simple schedules for the stream compositions.
When task structure demands, \ourmethod effectively learns task-specific schedules directly from demonstrations.
And \ourmethod even discovers the same schedules from multi-particle rollouts without specialized training.

\rebuttal{%
\textit{Limitations:} Multi-stream learning requires an object tracking module and mildly increases inference costs.
Future work might distill the streams into a single model.
Multi-task learning likely requires an additional module for task selection, which might also be investigated in future work, along with the inclusion of other modalities like haptic sensing.
}

\bibliographystyle{IEEEtran}
\bibliography{IEEEabrv,root}

\clearpage
\renewcommand{\baselinestretch}{1}
\setlength{\belowcaptionskip}{0pt}

\begin{strip}
\begin{center}
\vspace{-5ex}
\textbf{\LARGE \bf
MSG: Multi-Stream Generative Policies for Sample-Efficient\\\vspace{0.5ex}Robotic Manipulation} \\
\vspace{3ex}

\Large{\bf- Supplementary Material -}\\
\vspace{0.4cm}
\normalsize{Jan Ole von Hartz$^{*}$, Lukas Schweizer$^{*}$, Joschka Boedecker, and Abhinav Valada%
}
\end{center}
\end{strip}

\setcounter{section}{0}
\setcounter{equation}{0}
\setcounter{figure}{0}
\setcounter{table}{0}
\setcounter{page}{1}
\makeatletter

\renewcommand{\thesection}{S.\arabic{section}}
\renewcommand{\thesubsection}{S.\arabic{subsection}}
\renewcommand{\thetable}{S.\arabic{table}}
\renewcommand{\thefigure}{S.\arabic{figure}}

\section{Schedules}\label{sec:sched}
The logvar 2D and particle-based weighting schemes (which do not require a known object order) perform on par with the exponential schedule, except on \texttt{InsertOntoSquarePeg}
(and they outperform all baseline methods).
On \texttt{InsertOntoSquarePeg} they under-perform by about 12 percentage points (0.40 success rate versus 0.52) when using MCMC, and by 5 percentage points without MCMC.

\begin{figure}[h]
    \centering
    \includegraphics[width=\linewidth]{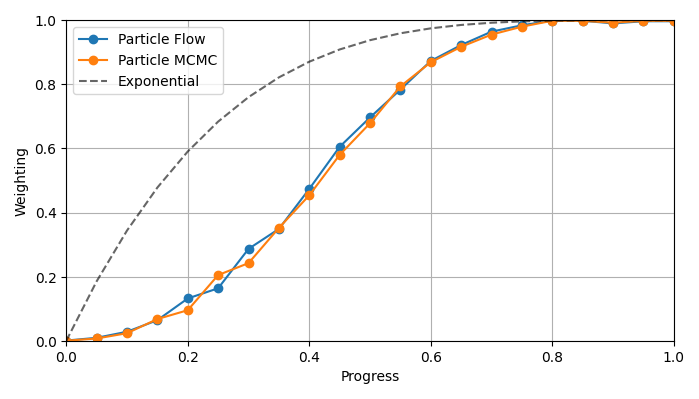}
    \caption{Particle-based weighting schedule versus exponential schedule on \texttt{InsertOntoSquarePeg}.}
    \label{fig:schedules}
\end{figure}

This task is an outlier because it has high precision demands while also having the largest task instance variation.
The square's pose is randomized in an area of 43.7x\SI{76.1}{cm}, rendering \texttt{InsertOntoSquarePeg} the hardest simulated task.
In contrast, \texttt{OpenDrawer} has the smallest area with 24.6x\SI{25.7}{cm} and is also one of the easiest tasks.
In general, the larger the task instance variation, the harder the task, because the policy has to generalize across a larger task space.
This is especially challenging given only 10 demonstrations.
The exponential schedule is advantageous in this case, because it weights the goal frame over the target frame earlier in the  trajectory.
This is shown in \figref{fig:schedules}.
The combination of little data with high variance makes the source frame less informative.
This also shows in the relative success rates of the single-frame policies in Tab.\ I (p.\ 7).
Here too, the single-frame End-Effector policy under-performs the Object-Frame policy more strongly on \texttt{InsertOntoSquarePeg} than on all other tasks.
We confirm this hypothesis by overwriting the particle-based schedule with the exponential schedule either in the first or second half of the trajectory.
As hypothesized, overwriting the first half of the schedule leads to the particle-based schedule closing the performance gap, whereas overwriting the second half does not.
(Overwriting the second half boosts policy success from 0.40 to 0.48, whereas overwriting the first half only achieves 0.41.)
These results indicate that high variance early in the trajectory (in conjunction with little data) is the driver of the observed performance gap.
The performance gap also vanishes if we slightly increase the number of demonstrations, again confirming our hypothesis.
In the manuscript this can be seen in Fig.\ 7b (p.\ 6).
From 20 demonstrations, the MSG Ensemble with the logvar-based schedule achieves near-identical policy success to the exponential schedule.

\begin{figure}[tb]
    \centering
    \includegraphics[width=0.14\textwidth,valign=t]{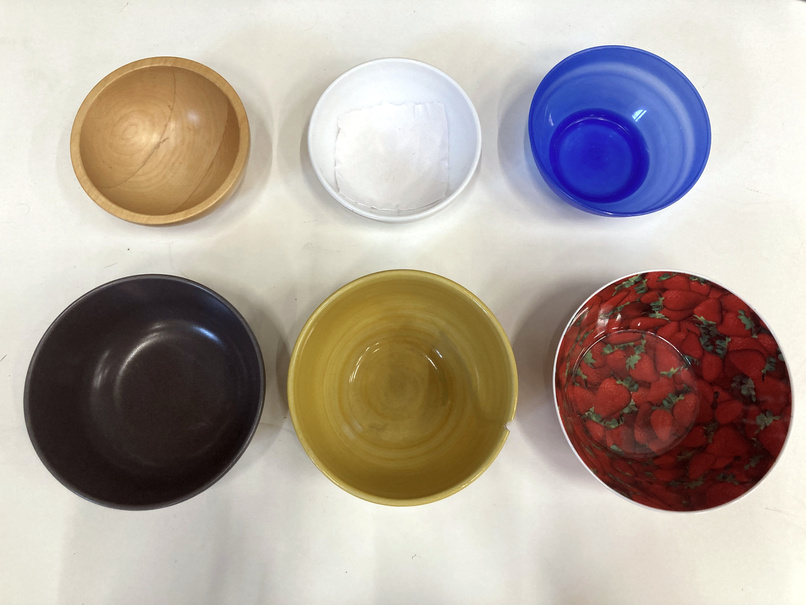}\hspace{0.1cm}%
    \includegraphics[width=0.105\textwidth,valign=t]{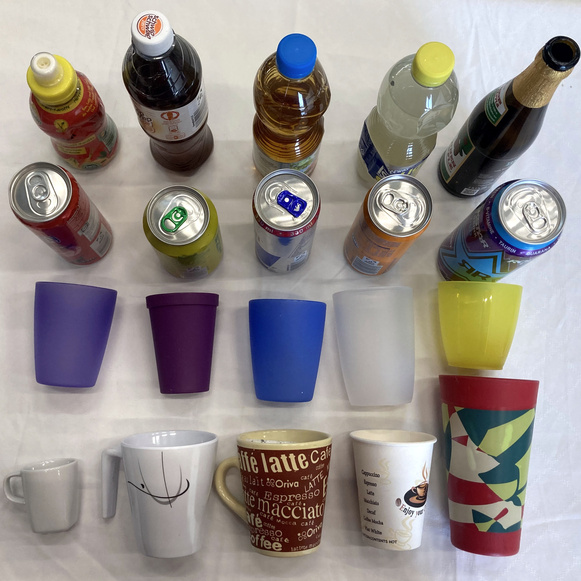}\hspace{0.1cm}%
    \includegraphics[width=0.105\textwidth,valign=t]{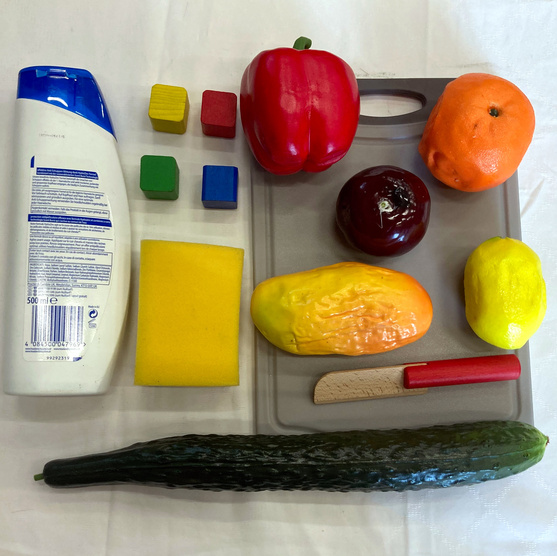}\hspace{0.1cm}%
    \includegraphics[width=0.105\textwidth,valign=t]{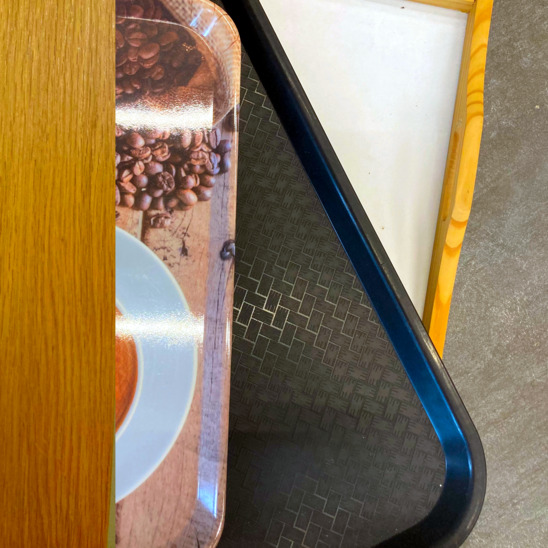}\\
    \vspace{0.1cm}
    {\footnotesize 
    \begin{tabular}{l c c c}
        \toprule
        & \multicolumn{3}{c}{\textbf{Variation}} \\
        \cmidrule(lr){2-4}
        \textbf{Task}        & {Object Instance} & {Clutter Objects} & {Environment} \\
        \midrule
        PickAndPlace    & 0.75     & 0.70        & 0.80             \\
        PourDrink       & 0.85     & 0.80        & 0.80             \\
        \bottomrule
    \end{tabular}
    }
    \caption{Objects and success rates for generalization study.
    \textit{Left:} object instances for both tasks, \textit{Center Right:} clutter objects, \textit{Right:} environments.
}\label{fig:obj_gen}
\end{figure}

\section{Keypoint Tracking}\label{sec:kp_track}
\subsection{DINO Encoder}
As noted in Sec.\ V.D (p.\ 8), the robustness of DINO-based keypoint detection has been studied in previous work~\cite{amir2021deep, vonhartz2024art,von2025unreasonable}.
In particular, von Hartz~\etal{}~\cite{vonhartz2024art,von2025unreasonable} have studied DINO-based keypoint detection in the context of multi-stream policy learning using Gaussian policies.
They found excellent robustness towards environment randomization, instance-level transfer, and clutter~\cite{vonhartz2024art,von2025unreasonable}.
As our supplementary video and Fig.\ 5 (p.\ 5) show, we have recreated the \texttt{PickAndPlace} and \texttt{PourDrink} tasks from~\cite{vonhartz2024art}, both of which are multi-object tasks.
Since we use the same keypoint-tracking strategy, their results should transfer to our setup.
To confirm this, we have conducted additional experiments, recreating the full set of object instance, clutter object, and environment randomizations shown in Fig.\ 7 (p.\ 6) in~\cite{vonhartz2024art}.
For each task, we evaluate 20 instances of object transfer per tasks (because we have up to 20 different object instances), as well as 10 instances of cluttered scenes and 10 instances with environment generalization.
The instances of cluttered scenes are comparable to the qualitative examples shown in our original supplementary video.
\figref{fig:obj_gen} shows the result of this generalization study.
As hypothesized, we observe comparable \emph{relative} policy success rates to von Hartz~\etal{}, with the differences accounted for by the different base policy success rates.
Due to space constraints in the main paper, we have now included these results on our project website, as well as adding a reference and short discussion in \rebuttal{Sec.\ V.D (p.8).}

\subsection{FoundationPose}
MSG is a modular approach.
End-to-end visuomotor policies are not only less interpretable and more expensive to train, but also expensive to update due to joint training of their perception and motor components.
In contrast, our approach improves in lockstep with advances in computer vision as the pose estimator can easily be swapped out.
We have evaluated MSG with FoundationPose on the \texttt{PickAndPlace} task as well, finding near-identical policy performance to the DINO-variant (task success of 0.80).
Overall, we have observed similar tracking performance and thus policy success.
Other works similarly report good results using both tracking systems~\cite{von2025unreasonable}.
In the end, we have opted for using DINO in our experiments for four main reasons.
First, while FoundationPose does not necessarily need a CAD model of the object to track, it \emph{does} need an object mesh.
In the absence of a CAD model, the mesh is reconstructed by scanning the object with an RGB-D camera beforehand and applying BundleSDF~\cite{wen2024foundationpose}.
This scan and the following mesh reconstruction do not only take some time, but also do not work equally well for all objects.
For example, reconstructing the banana for \texttt{PickAndPlace} worked well, whereas reconstructing and tracking the drawer (or only its handle) did not.
There, drawer was simply too big to scan and track well, whereas the handle has few features and cannot be scanned well due being attached to the drawer.
Second, FoundationPose's peak GPU memory usage is quite high.
Whereas the used DINO encoder only uses around \si{1}{GB} of GPU memory (for images of 480x640px), FoundationPose can peak well above \si{10}{GB}, depending on the object.
Third, using FoundationPose has a higher inference latency to establish the initial pose.

\begin{table*}[tb]
    \centering
    \begin{tabular}{l c c c c}
        \toprule
        \textbf{Method} & \textbf{Raw Wall-Clock Time [ms]} & \textbf{Normalized Wall-Clock Time [ms]} & \textbf{Effective Frequency [Hz]} \\
        \midrule
        Vanilla Flow Matching & \phantom{0}89 & 11.1 & 90.1\\
        Flow Ensemble & 100 & 12.5 & 80.0 \\
        Flow Composition & 150 & 18.8 & 53.2\\
        MCMC & 580 & 72.5 & 13.8\\
        \bottomrule
    \end{tabular}
    \caption{Inference time of policies.
    The normalized wall-clock time is calculated by dividing the raw wall-clock time by the action horizon (which is 8).
    The effective frequency is the inverse of the normalized wall-clock time.
    In our experiments, a frequency of \SI{20}{Hz} constitutes real-time capability of the policy.
    }
    \label{tab:inf_time}
\end{table*}

\section{Inference Cost}\label{sec:inf_cost}
Following standard practice, we predict a sequence of 16 actions at a time (prediction horizon), of which 8 are then executed (action horizon)~\cite{chi2023diffusionpolicy, chisari2024flowmatch, wang2024inference}.
\tabref{tab:inf_time} therefore reports both the raw wall-clock time per prediction, as well as the wall-clock time normalized by the \emph{action} horizon.
It also reports the effective prediction frequency as the inverse of the normalized wall-clock time.
As we control the robot with a frequency of \SI{20}{Hz}, the same frequency constitutes real-time control.

Overall, the multi-stream prediction itself adds only a modest overhead as the parallel integration of two streams is not particularly costly.
The Ensemble increases the inference latency by about 12\%, whereas the Flow Composition adds around 69\%.
The large discrepancy between Ensemble and Flow Composition stems primarily from the repeated intermediate frame transformations required by the latter that we perform using non-optimized Python code.
As both models run well within real-time, we have not optimized the code for speed.
The optional MCMC scheme is computationally significantly more expensive, increasing the inference cost of Flow Composition roughly four times.
This finding is inline with prior work~\cite{wang2024inference} and stems from the fact that MCMC runs four times as many flow integration steps.
Note that this is a constant factor, hence not influencing the asymptotic complexity of the algorithm.
Other factors such as code optimizations, policy architecture, and improving hardware should therefore be able to compensate for this constant factor.
Even now, with an effective inference frequency of \SI{13.8}{Hz}, MCMC is already close to running in real-time.
Code optimizations, tuning the number of inference steps, or tuning the prediction horizon should allow MCMC to be real-time capable in practice.
Note further that the advantage of MCMC does not simply stem from using more compute.
As we show in Tab.\ III (p.\ 7), simply matching the number of inference steps used by MCMC without performing the MCMC-scheme itself does not achieve the same policy performance.
If compute is limited, our Ensemble approach offers a computationally cheap alternative, running at \SI{80}{Hz}, and still outperforming all baseline models.

\end{document}